\documentclass[pdflatex,sn-nature]{sn-jnl}
\usepackage{graphicx}
\usepackage{float}
\usepackage{amsmath,amssymb,amsfonts}
\usepackage{booktabs,multirow,array}
\usepackage{url}
\begin{document}
\title[Scientific feedback and generative discovery]{Scientific feedback reveals productive regimes beyond training conventions}
\author[1]{Yoon Pyo Lee}
\author[1]{Samrendra Roy}
\author[1,3,4]{Souvik Chakraborty}
\author[1,2]{Syed Bahauddin Alam\textsuperscript{*}}\email{alams@illinois.edu}

\affil[1]{\orgdiv{Department of Nuclear, Plasma, and Radiological Engineering}, \orgname{The Grainger College of Engineering, University of Illinois Urbana-Champaign}, \orgaddress{\city{Urbana}, \state{IL}, \country{USA}}}
\affil[2]{\orgdiv{National Center for Supercomputing Applications}, \orgname{University of Illinois Urbana-Champaign}, \orgaddress{\city{Urbana}, \state{IL}, \country{USA}}}
\affil[3]{\orgdiv{Department of Applied Mechanics}, \orgname{Indian Institute of Technology Delhi}, \orgaddress{\city{New Delhi}, \country{India}}}
\affil[4]{\orgdiv{Yardi School of Artificial Intelligence}, \orgname{Indian Institute of Technology Delhi}, \orgaddress{\city{New Delhi}, \country{India}}}

\abstract{
Human-imposed design conventions can restrict the alternatives represented in scientific training data and shape the search preferences learned by generative models. We show that scientific performance feedback can redirect the learned search distribution toward productive regions beyond those conventions, without explicitly rewarding departures from them. We test this process in reactor lattice design with simulated feedback and chemical reaction-condition selection with measured yields, using a pretrained sequence policy and a small non-language policy, respectively. In both settings, adaptation improves scientific performance while increasing generation of choices excluded from the corresponding supervised data. In measured reactions, probability on task-specific ligand choices excluded from training rises from 0.29 to 0.70--0.71 as expected yield error decreases. Controls in the lattice setting show that the productive region identified during adaptation provides reusable information to random search, while aligned generation produces useful candidates more frequently than informed random sampling. Removing the target penalty redirects four of five matched runs, linking the direction of adaptation to the scientific objective. Together, these results show how scientific feedback reshapes inherited search preferences and reveals productive alternatives beyond training conventions, with attribution controls distinguishing learned redistribution from candidate availability and reusable search information.
}
\maketitle

\section{Introduction}\label{sec:intro}

Scientific design begins with decisions about what may change. Experts choose material inventories, geometric symmetries, topology classes and admissible parameter ranges before optimization begins \cite{molesky2018inverse,sanchez2018inverse,zheng2023unifying}. These choices make large design problems tractable and incorporate accumulated knowledge. They also determine which alternatives are evaluated and recorded. Training corpora inherit this selection, so a model learns from examples already shaped by human assumptions. Human experimental choices can bias scientific datasets, and neural models can acquire regularities that transfer poorly to changed tasks \cite{jia2019anthropogenic,geirhos2020shortcut}. A convention repeated in every training example may consequently be learned as though it were an intrinsic property of the physical system. When scientific requirements change, useful solutions may lie beyond that convention. This makes reconsidering inherited design choices an important problem for scientific machine learning. Here we study this problem in reactor lattice design and chemical reaction-condition selection, using simulated feedback and previously measured reaction yields, respectively.

Machine learning has substantially extended the tools available for scientific design. Learned structure--property maps, inverse models and active learning support the search for architected materials with prescribed mechanical responses \cite{bastek2022inverting,peng2023machine,zheng2023unifying,maurizi2025designing}. These advances improve prediction and search within specified representations and design variables. The choice of those variables remains consequential. Holding an inventory or topology fixed simplifies a search while placing other regimes outside its evaluated space. For a learned generator given access to broader choices, the corresponding challenge is to use them productively when its supervised examples consistently favor the original convention.

Generative models offer a direct way to study this problem because they learn distributions over complete candidate structures. Molecular and crystal generators connect structural representations with desired properties \cite{elton2019deep,bilodeau2022generative,antunes2024crystal,xiao2023invertible}. Graph-based approaches combine imitation learning, reinforcement learning and tree search to generate metamaterials with prescribed nonlinear responses and geometric constraints \cite{maurizi2025designing}. Algebraic language models encode shell geometries as mathematical sentences and use diffusion transformers for inverse design, including mechanical responses beyond their training domains \cite{zheng2026algebraic}. In quantum experiments, language models generate programs that encode reusable construction rules and extrapolate to larger systems \cite{arlt2026meta}. For sequence policies, domain-adaptive pre-training and instruction fine-tuning provide ways to adapt generation to available examples and requested specifications \cite{gururangan2020don,wei2021finetuned,ouyang2022training}. The resulting prior reflects both useful scientific relationships and recurring choices in the examples. When one feature takes the same value throughout training, the corpus supplies no direct comparison with alternative values. Scientific evaluation can provide that comparison by scoring newly generated candidates and guiding subsequent updates.

This integration of generation and evaluation is becoming central to scientific discovery workflows \cite{wang2023scientific,chen2026xray,shi2026matbrain}. Reward-guided methods have advanced algorithmic discovery \cite{fawzi2022discovering}. In molecular design, memory-based reinforcement learning improves sample efficiency, with analyses also examining out-of-support optimization, generation likelihoods and sampled property distributions \cite{guo2026sample}. In protein engineering, reinforcement learning guided by a fitness model trained on single-mutation data has identified experimentally validated multi-mutant sequences \cite{sun2025accelerating}. Experimental outcomes can also reshape language-model representations to guide Bayesian optimization across scientific tasks \cite{rankovic2025large}. In peptide design, adapting a generative representation during Bayesian optimization supports improved antimicrobial activity under explicit template-similarity constraints \cite{torres2026generative}. Reinforcement learning also guides crystal generation through explicit rewards for stability, diversity and novelty \cite{park2026guiding}. Earlier work examines targeted molecular generation when examples with the desired properties are scarce or absent from training \cite{mokaya2023testing}.

These studies establish that learned scientific search can generate useful alternatives beyond its training examples and that adaptation can change the sampled distribution. We ask whether scientific performance feedback can redirect generation toward productive regions beyond a convention consistently represented in task-specific supervised data. The policy receives no explicit reward for changing the convention-defining feature or for novelty. The permitted candidate space remains fixed, and we measure how adaptation redistributes generation probability within it. A successful candidate alone does not establish how often useful candidates occur under reference sampling, how adaptation changes their generation frequency, or what information about the productive region another search can reuse.

To evaluate this movement beyond the training convention, we develop an experimental attribution framework that distinguishes candidate availability, learned redistribution and information value. Availability asks where useful candidates occur and how often reference sampling encounters them. Redistribution asks how adaptation changes the policy and the frequency of useful unselected generations. Information value asks what a simpler search can recover when given the productive region identified during adaptation. An objective intervention tests whether the direction of adaptation follows the rewarded scientific quantity. Motivated by calls for component attribution and comparisons with simpler workflows \cite{nmi2026transparency}, we use these controls to evaluate both the adapted generator and the information that another search procedure can reuse. This makes the contribution of learned adaptation testable beyond the quality of its single best output.

The lattice and reaction experiments use a pretrained sequence policy and a small categorical policy, respectively. The sequence policy learns from reactor lattices containing exactly 16 absorber rods, then receives OpenMC feedback while generating layouts with variable inventory \cite{romano2015openmc}. The categorical policy learns from a reaction catalog restricted to one ligand per aryl-halide task and subsequently receives feedback from the full measured catalog \cite{ahneman2018predicting}. Exact enumeration measures its candidate probabilities and expected yield error. Online variants of DPO and GRPO provide pairwise and group-relative updates in both settings \cite{rafailov2023direct,shao2024deepseekmath}. These experiments combine structured generation under noisy physical evaluation with a non-language architecture whose distribution can be evaluated completely.

Across both settings, scientific feedback improves candidate quality and directs generation beyond the corpus convention. In the lattice experiment, both update rules identify productive inventories of 28--35 absorbers. Objective removal redirects the adaptation, unselected generation measures its concentration on useful candidates, and informed random search establishes the value of the identified region. The chemistry experiment quantifies reward-directed redistribution and expected performance exactly using measured reaction yields and a small non-language policy. These findings connect adaptation beyond training conventions to productive generation and reusable design information. Additional initialization experiments characterize the learned policies under different training histories.

\section{Results}\label{sec:results}

\begin{figure}[H]
\centering
\includegraphics[width=0.98\textwidth]{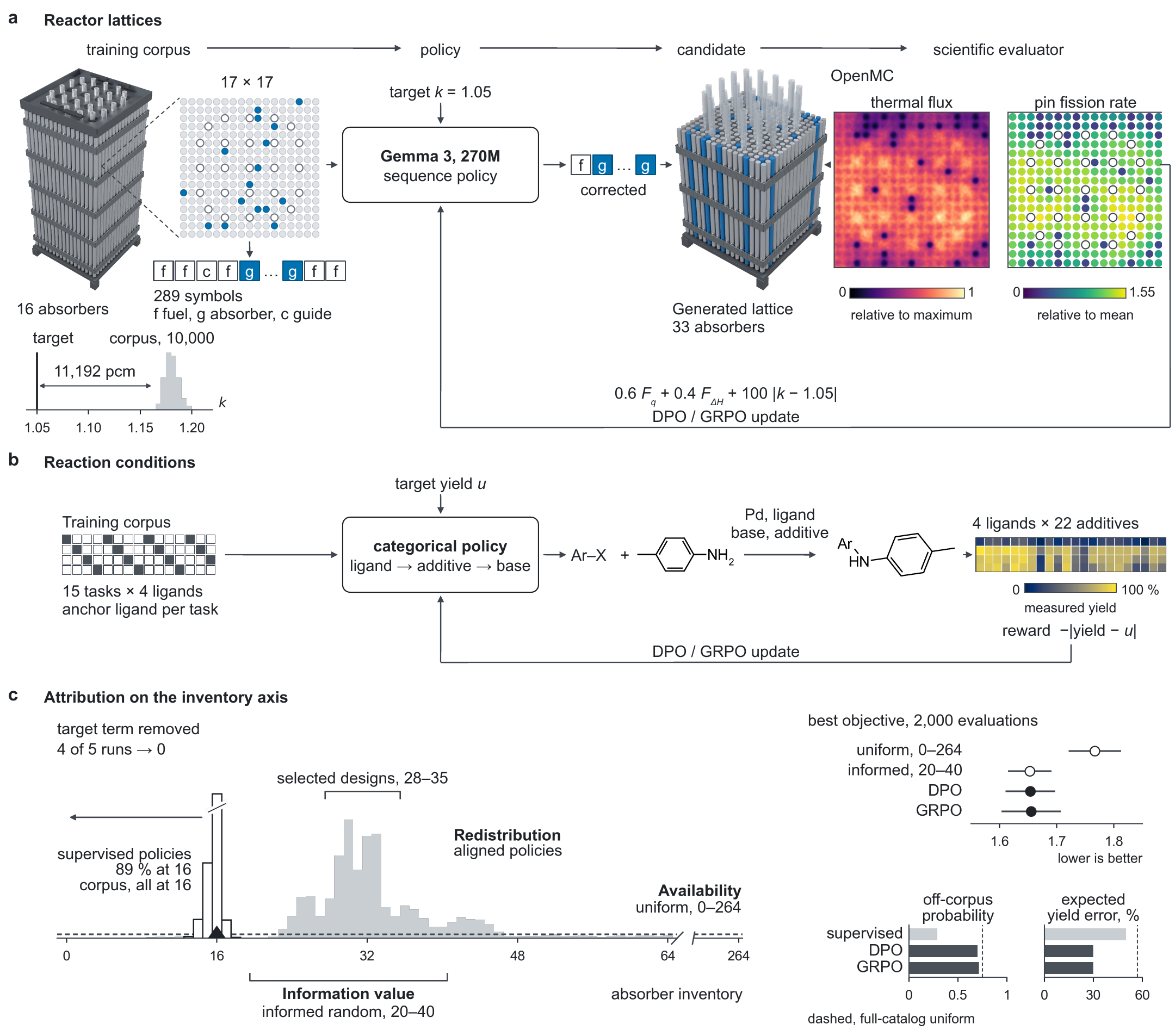}
\caption{\textbf{Scientific feedback beyond training conventions in two implementations and the attribution of the adaptation.} \textbf{a} Reactor lattices. One of 10{,}000 training lattices, all with 16 absorber rods, is serialized into 289 symbols for fuel (\texttt{f}), absorber (\texttt{g}) and guide channel (\texttt{c}). The histogram gives the corpus multiplication factor, whose nearest lattice lies 11{,}192 pcm above the target of 1.05. Gemma 3 with 270 million parameters, trained on this corpus, generates text for the prompt $k=1.05$ that is corrected and evaluated by OpenMC. The generated lattice is the selected DPO design with 33 absorbers. Maps give the thermal neutron flux ($E<0.625$ eV) relative to its maximum and the pin fission rate relative to the cell mean ($k=1.0492$). \textbf{b} Reaction conditions. Filled cells mark the one anchor ligand per aryl-halide task kept in the supervised corpus from 3{,}955 measured Buchwald--Hartwig reactions. A categorical policy selects ligand, additive and base among measured rows for the coupling with p-toluidine. Colours give measured yields for 3-bromopyridine with MTBD over the four ligands (rows, XPhos, t-BuXPhos, t-BuBrettPhos and AdBrettPhos) and 22 additives, and the reward is the negative absolute yield error. \textbf{c} Attribution on the inventory axis. Outlined bars give the inventory of 5{,}000 corrected outputs from the five supervised policies before alignment (Methods), 88.6\% at exactly 16 and none above 18. The triangle marks the corpus and the tallest bar is clipped. Grey bars give the inventory of 10{,}000 corrected, unselected outputs from the ten aligned checkpoints, 5.2\% of them above 64. The dashed line is the inventory-uniform density over 0--264 (availability). The upper bracket marks 28--35, the inventories of the ten selected layouts (redistribution). The arrow marks objective removal, after which four of five matched runs select zero absorbers. The lower bracket marks the interval 20--40 supplied to informed random search (information value). Right, best objectives after 2{,}000 evaluations (five-run mean and standard deviation) and the exact off-corpus ligand probability and expected yield error at requested yield 90 (five-seed means), with the full-catalog uniform sampler dashed.}
\label{fig:concept}
\end{figure}

We evaluate how scientific feedback redirects generation beyond training conventions in reactor lattice design and chemical reaction-condition selection. The representations and feedback loops for the two settings are shown in Fig.~\ref{fig:concept}a,b. Figure~\ref{fig:concept}c summarizes the distributional measurements in both settings and the attribution controls in the lattice experiment.

\subsection{Scientific feedback discovers a productive regime beyond the corpus}\label{sec:reactor}

The lattice experiment tests whether a policy can identify useful values of a design variable held constant throughout its supervised corpus. ReactorGen represents a $17\times17$ lattice with 289 cell symbols for ordinary fuel, gadolinium-bearing absorber fuel and guide channels. A Gemma 3 backbone with 270 million parameters \cite{kamath2025gemma,vaswani2017attention} undergoes continued pre-training on unconditional layouts and supervised fine-tuning on layouts paired with evaluated scientific quantities \cite{gururangan2020don,ouyang2022training,wei2021finetuned}. All 10{,}000 source configurations contain 16 absorbers. Online generation permits fuel or absorber choices at 264 positions and retains 25 fixed guide positions.

Each generated candidate receives a fresh OpenMC evaluation. The objective combines multiplication-factor error with power peaking,
\begin{equation}
 \mathcal L_{\mathrm{fit}}(z;u)=0.6F_q(z)+0.4F_{\Delta H}(z)+100|k(z)-u|,
 \label{eq:fitness}
\end{equation}
where $u=1.05$ and lower values are better. The reflective two-dimensional model gives $k_\infty$, denoted by $k$ here. The source corpus lies far from the requested target. Its nearest configuration is more than 11{,}000 pcm away, where 1 pcm is $10^{-5}$ in $k$. The reward supplies the target and power-distribution objective while absorber inventory remains a choice made by the policy.

\begin{figure}[H]
\centering
\includegraphics[width=\textwidth]{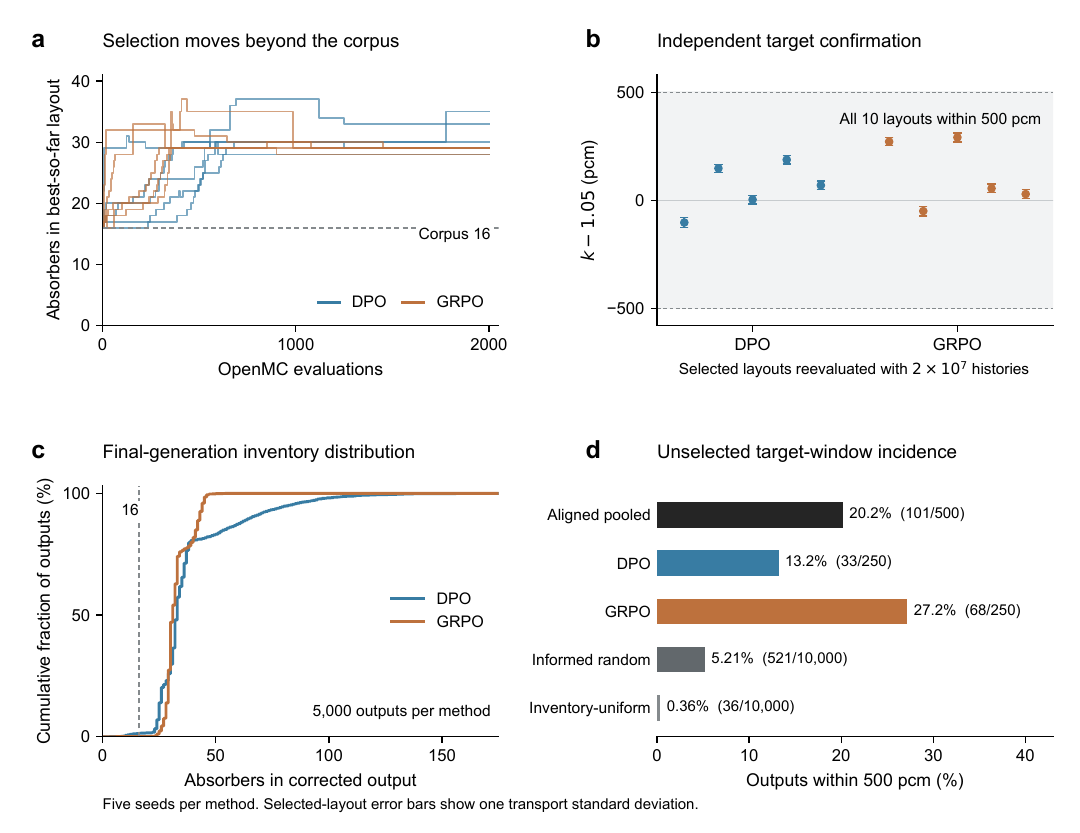}
\caption{\textbf{Scientific feedback identifies and populates a productive design regime.} \textbf{a} Inventory of best-so-far layouts during DPO and GRPO alignment, with one line per training seed. The horizontal reference marks the 16-absorber corpus convention. GRPO is displayed at group-completion times. \textbf{b} Signed deviations from the requested multiplication factor 1.05 after independent reevaluation of all ten selected layouts. Each evaluation uses 20 million active histories. Error bars show one Monte Carlo standard deviation. Every layout remains within 500 pcm of the target. \textbf{c} Empirical cumulative distributions of inventory in 5{,}000 corrected, unselected outputs per method, with 1{,}000 per checkpoint at temperature 1.0. \textbf{d} Target-window incidence among 500 randomly selected corrected outputs, with 50 per checkpoint, and 10{,}000 candidates from each random sampler. DPO contributes 33/250 hits and GRPO 68/250, giving 101/500 in the pooled aligned sample. Informed random sampling over inventories 20--40 gives 521/10{,}000 hits. Inventory-uniform sampling over 0--264 gives 36/10{,}000. Methods specifies the sampling procedures.}
\label{fig:discovery}
\end{figure}

Both online update rules identify a productive inventory regime absent from the corpus. Across five seeds and 2{,}000 evaluations per run, the best DPO layouts contain 28--35 absorbers and the best GRPO layouts contain 28--29. Mean best objectives reach $1.65\pm0.04$ and $1.66\pm0.05$, respectively, with uncertainty reported as sample standard deviation across seeds. Figure~\ref{fig:discovery}a follows the inventory of each method's improving designs. The adaptation uses only scientific performance as feedback, with no inventory, novelty or diversity term in the objective.

Both initialization histories support productive adaptation, with lower across-seed objective dispersion after CPT$+$SFT. Supplementary Information reports the complete initialization comparison.

Independent high-statistics evaluation confirms target attainment for all ten selected policy layouts (Fig.~\ref{fig:discovery}b). Their online target errors are $44\pm36$ pcm for DPO and $40\pm28$ pcm for GRPO. Reevaluation with 20 million active histories per layout gives $102\pm71$ and $139\pm130$ pcm. Transport uncertainty decreases from approximately 150 pcm online to approximately 20 pcm in this check. Every reevaluated layout remains inside the 500-pcm target window. The corresponding reevaluated composite objectives are $1.515\pm0.062$ for DPO and $1.551\pm0.134$ for GRPO. The lower mean composites reflect a decrease in the noise-sensitive maximum pin-power ratio $F_q$ across all ten layouts.

Unselected generation makes the shift beyond the training convention visible before and after alignment. Using the same generation and correction procedure, we draw 1{,}000 outputs from each of the five seed-specific supervised CPT$+$SFT policies. Among these 5{,}000 corrected outputs, 88.6\% retain exactly 16 absorbers and none contains more than 18 (Fig.~\ref{fig:concept}c). After alignment, we draw 1{,}000 outputs from each of the ten final single-target CPT$+$SFT checkpoints. Across this 10{,}000-output pool, corrected inventory is $35.8\pm14.3$ absorbers, reported as the mean and standard deviation across outputs.

We randomly select 50 corrected outputs per aligned checkpoint for transport evaluation, without fitness-based selection. Of these 500 candidates, 101 reach the target window, giving 20.2\% incidence. No source-corpus configuration lies in this window. Together with the selected layouts, these unselected samples connect the change in the generation distribution to a productive higher-inventory regime (Fig.~\ref{fig:discovery}c,d).

DPO contributes 33 target-window candidates among 250 evaluated samples and GRPO contributes 68 among 250, giving rates of 13.2\% and 27.2\%. Their similar best objectives coexist with different frequencies of productive unselected generation. This distinction makes the policy distribution an informative endpoint alongside best-of-budget quality. The evaluated generator includes deterministic length and guide-coordinate correction. Every absorber token in the corrected layouts originates in model generation. Methods reports the complete correction procedure and output-format measurements.

\subsection{Objective interventions and information handoff identify the contributions of adaptation}\label{sec:controls}

The attribution framework connects three measurements. Inventory-uniform sampling characterizes candidate availability in the supplied space. Policy samples measure how frequently learned generation produces useful candidates. Informed sampling measures the value of passing an identified region to a simpler search. Objective removal then tests which scientific reward directs the adaptation.

\begin{figure}[H]
\centering
\includegraphics[width=0.92\textwidth]{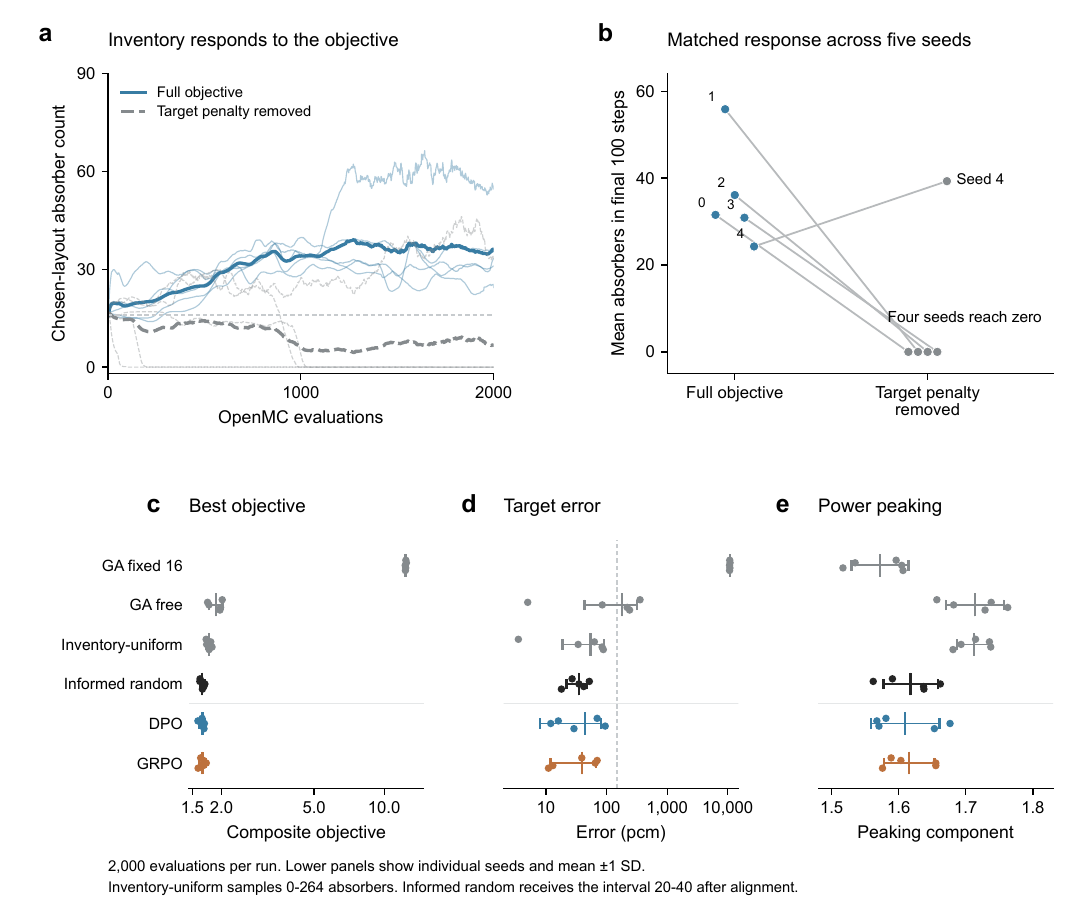}
\caption{\textbf{Objective dependence and the value of inventory information.} \textbf{a} Chosen-layout inventory trajectories for full-objective DPO and matched runs with the target penalty removed. Thin lines show seeds and thick lines show the five-seed mean after a 25-step rolling mean. \textbf{b} Final-100-step mean inventories for each matched seed. Four removed-penalty runs reach zero inventory and the fifth averages 39.29. \textbf{c--e} Best-of-budget composite objective, absolute target error and peaking-only component from five independent runs of 2{,}000 online evaluations per method. Points show seeds and bars show means with sample standard deviations. Inventory-uniform random search samples count over 0--264. Informed random search receives the interval [20,40] identified after alignment and recovers comparable objective quality. The dashed reference in \textbf{d} marks the approximate online transport uncertainty of 150 pcm. The fixed-inventory GA uses 16 absorbers.}
\label{fig:controls}
\end{figure}

\textbf{Objective-directed adaptation.} Five matched DPO runs remove the multiplication-factor penalty from Equation~\ref{eq:fitness}. The prompt, initialization, sampling temperature, correction, transport settings and online budget remain fixed. With the resulting objective $0.6F_q+0.4F_{\Delta H}$, four seeds select exclusively zero-absorber candidates over their final 100 steps. The fifth retains a higher inventory, averaging 39.29 absorbers over the same interval. Across the five seed-level means, inventory is $7.86\pm17.57$ after penalty removal and $35.73\pm12.02$ under the full objective. Figure~\ref{fig:controls} shows the paired responses and the retained exception. Changing the scientific target term therefore changes the predominant direction of inventory adaptation.

\textbf{Candidate occurrence and probability concentration.} The uninformed sampler draws absorber count uniformly from integers 0--264 and positions uniformly conditional on that count. Five runs of 2{,}000 evaluations produce 36 target-window candidates among 10{,}000 samples. Per-run counts are 6, 11, 10, 5 and 4. All 36 successful candidates contain 28--35 absorbers, and 281 sampled layouts occupy this interval. The productive region is reached in every random-search run.

Corrected unselected generations from the final aligned checkpoints attain 20.2\% target-window incidence, compared with 0.36\% for this inventory-uniform sampler (Fig.~\ref{fig:discovery}d). The approximately 56-fold difference shows a substantial concentration of useful candidates under the learned generation procedure. Informed random sampling gives 521 target-window candidates among 10{,}000 samples, or 5.21\%. The aligned incidence is approximately 3.88 times this informed rate. Within inventories 28--35, target-window incidence is 83/282 (29.4\%) for aligned outputs, 521/3{,}830 (13.6\%) for informed random samples and 36/281 (12.8\%) for inventory-uniform samples. Supplementary Information reports the counts for each procedure.

\textbf{Information passed to a simpler search.} Informed random search receives the interval [20,40] selected after observing the alignment results. This coarse bracket contains the 28--35 inventories of the selected aligned layouts. The sampler draws inventory uniformly within the bracket, then samples positions uniformly at that count. Under the same 2{,}000-evaluation budget, its mean best objective is $1.65\pm0.04$. The aligned policies attain $1.65\pm0.04$ for DPO and $1.66\pm0.05$ for GRPO. Exact two-sided Mann--Whitney comparisons give $U=12$ and $p=1.0$ with five runs per group. Once supplied with the inventory interval, a simple placement rule recovers comparable best design quality.

This information handoff establishes the productive region as a compact design rule with value beyond the original generator. Inventory information allows simple random placement to recover comparable best-of-budget quality, while the aligned policies retain a higher frequency of target-reaching unselected candidates. Best-of-budget quality alone would therefore miss differences in the frequency of useful generation.

\begin{table}[htbp]
\centering
\caption{Best-of-budget objectives across five runs of 2{,}000 online evaluations. Values are means $\pm$ sample standard deviations. The aligned policies also use a shared 10{,}000-evaluation training corpus. Informed random search receives the interval identified from alignment.}
\label{tab:controls}
\begin{tabular}{llc}
\toprule
Procedure & Absorber-count access & Best objective\\
\midrule
Constrained GA & Fixed at 16 & $12.38\pm0.06$\\
Unconstrained GA & 0--264 & $1.90\pm0.13$\\
Inventory-uniform random & 0--264 & $1.767\pm0.045$\\
Informed random & Supplied interval 20--40 & $1.65\pm0.04$\\
DPO policy & 0--264 & $1.65\pm0.04$\\
GRPO policy & 0--264 & $1.66\pm0.05$\\
\bottomrule
\end{tabular}
\end{table}

The evolutionary comparisons measure selected-design quality under the same online budget. The unconstrained genetic algorithm receives the same objective, action space, simulator and online evaluation budget. DPO and GRPO improve its mean best objective by 12.7\% and 12.6\%, respectively, using unrounded values. Exact two-sided Mann--Whitney comparisons give $p=0.0079$ for each policy. Inventory-uniform random search also improves on the GA mean, reaching $1.767\pm0.045$. The fixed-16 GA attains $12.38\pm0.06$ and remains far from the target (Table~\ref{tab:controls} and Fig.~\ref{fig:controls}).

\subsection{Measured reaction rewards redirect generation beyond task-specific training choices}\label{sec:empirical}

The reaction setting complements the lattice experiment by permitting exact measurement of how generation probability and expected target error change under feedback from measured yields. A small non-language, masked categorical autoregressive policy with hidden size 48 selects ligand, additive and base for a requested reaction target. Its evaluator is the published Buchwald--Hartwig dataset of Ahneman et al.\ \cite{ahneman2018predicting}. The \texttt{Plates1-3} worksheet contains 3{,}955 measured combinations of 15 aryl-halide tasks, four ligands, 22 additives and three bases. A prefix mask admits measured combinations, so each reward is retrieved from a previously measured experimental yield.

Supervised training uses one anchor ligand per task, assigned by an outcome-independent task-index rule. All four ligand identities occur globally, while three quarters of task--ligand pairings are absent from the restricted corpus. Online generation permits the full measured catalog. Reward is the negative absolute difference between measured and requested yield. Ligand identity, novelty and diversity have no separate reward terms. Exact catalog enumeration measures expected target error and the probability assigned to task-specific ligand choices excluded from training.

\begin{figure}[H]
\centering
\includegraphics[width=\textwidth]{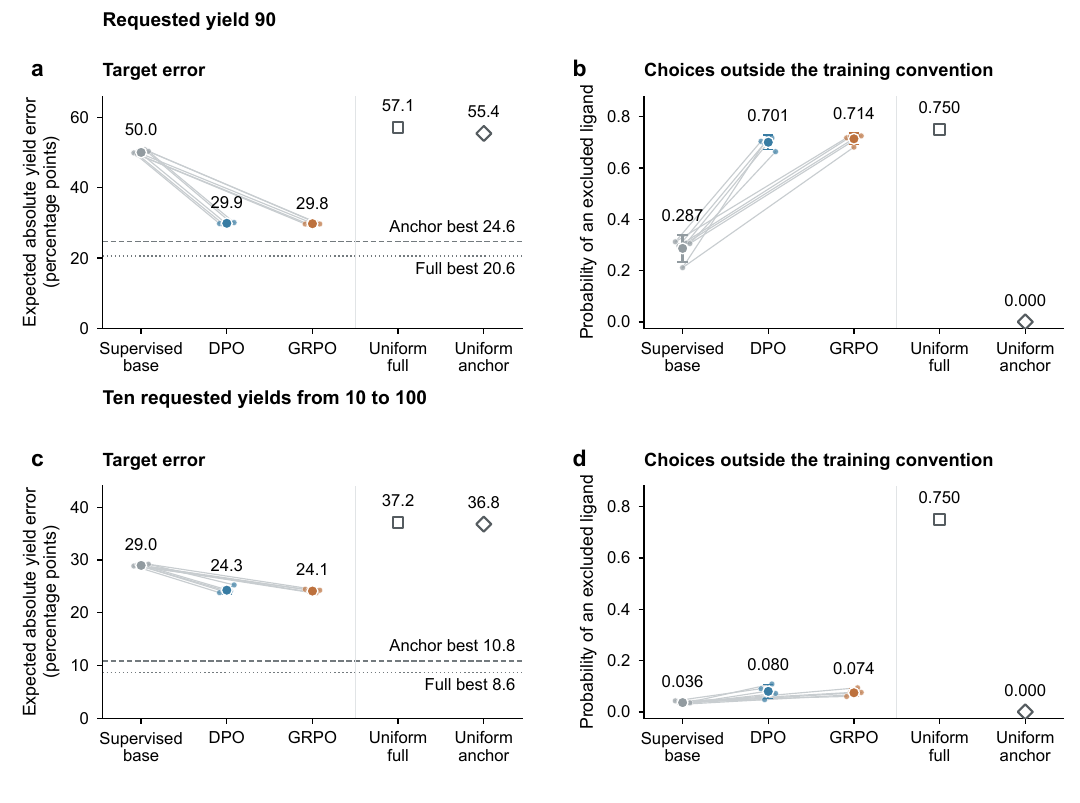}
\caption{\textbf{Scientific rewards select useful reactions beyond task-specific training choices.} \textbf{a,b} Expected absolute yield error and off-corpus ligand probability for requested yield 90. \textbf{c,d} Corresponding measurements for ten balanced targets from 10 to 100. The supervised base, DPO and GRPO use CPT$+$SFT initialization. Gray lines connect each of five seeds to its aligned descendants. Large markers and intervals show means and two-sided 95\% Student-$t$ intervals across seeds. Uniform full-catalog and anchor-catalog samplers provide exact distributional comparisons. Error panels also show the attainable minima in the full and anchor catalogs. All within-policy quantities are exact catalog expectations.}
\label{fig:empirical}
\end{figure}

For requested yield 90, CPT$+$SFT initialization gives expected absolute error 50.0 percentage points and off-corpus ligand probability 0.29. After 2{,}000 reward evaluations, DPO and GRPO reduce error to 29.9 and 29.8 while increasing off-corpus probability to 0.70 and 0.71 (Fig.~\ref{fig:empirical}a,b). Every paired seed shows both changes under both update rules. The SFT-only arm reproduces the same directions. Its higher starting off-corpus probability of approximately 0.67 rises to approximately 0.73 under either rule. Scientific feedback thus changes both the quality of the generated reactions and their use of alternatives excluded from task-specific training.

Uniform catalog sampling clarifies the selectivity of this redistribution. A full-catalog uniform sampler assigns approximately 0.75 probability to off-corpus ligand choices and has expected error 57.1 percentage points. Uniform sampling of anchor-ligand rows has error 55.4. The aligned policies attain much lower error while allocating slightly less off-corpus probability than the full uniform sampler. Their advantage comes from concentrating on useful reactions within the available choices. The lowest task-averaged attainable error is 24.6 within the anchor catalog and 20.6 over the full catalog. The excluded choices therefore also contain reactions with better attainable performance.

Multi-target training produces a smaller redistribution from a different supervised starting distribution. For ten balanced yields from 10 to 100, expected error decreases from 29.0 to 24.3 for DPO and 24.1 for GRPO. Off-corpus probability increases from 0.04 to 0.08 and 0.07 (Fig.~\ref{fig:empirical}c,d). Full- and anchor-catalog uniform sampling give errors 37.2 and 36.8. Every paired seed again combines reduced error with increased off-corpus probability. Both SFT example selection and online rewards use the broader target schedule. Within each training setting, the paired comparisons show that scientific feedback improves expected performance while increasing use of excluded ligand choices. Supplementary Information reports both initialization histories and all paired changes.

\section{Discussion}\label{sec:discussion}

Scientific feedback reveals productive alternatives to conventions inherited from a training corpus. In the lattice experiment, both reward-driven update rules identify useful inventories excluded from every supervised example. The final policies generate target-reaching designs frequently, and a small categorical model reproduces the coupled improvement in quality and off-corpus probability using measured reaction rewards. In both settings, performance feedback increases the use of choices excluded by the supervised training convention while improving candidate quality.

The attribution framework separates candidate availability, learned redistribution and information value. Broad random sampling reaches productive candidates in every run. The aligned generation procedure produces them substantially more frequently, including when compared with a sampler already supplied with inventory information. Passing the identified inventory bracket to random search recovers comparable best design quality. The region thus provides a reusable design rule, while the generated distribution captures an additional advantage in how often useful candidates are produced. Objective removal connects the direction of adaptation to the rewarded scientific quantity. Taken together, these experiments turn the attribution of discovery into a set of observable comparisons among candidate occurrence, policy concentration and information reuse.

The chemistry experiment makes the distributional result especially clear. Full-catalog uniform sampling places substantial mass beyond the training convention, yet its expected target error remains high. Alignment directs probability toward useful combinations and improves expected performance under both target schedules. Exact enumeration links these changes at the policy level, while paired seeds establish their consistency. The recurrence in a small non-language architecture broadens the result beyond pretrained sequence models and simulator-specific feedback.

The two evaluators provide complementary views of this process. A finite experimental catalog supports exact probability and expected-error calculations. Stochastic transport tests long structured generation under noisy scientific feedback. Independent reevaluation confirms the target attainment of the selected lattice designs, and computational accounting separates the online search budget from corpus construction and assessment. The detailed assembly physics, selected layouts and depletion behavior remain available in Supplementary Information. A supplementary prompt sweep on earlier archived checkpoints examines how generated inventory responds to changing requested targets.

The experimental logic can guide evaluation when a feature held fixed in supervised data is allowed to vary during policy adaptation. It separates improvement in unselected generation, dependence on the rewarded objective and information reusable by another search procedure. For generative modeling, reinforcement learning and learned search, these comparisons help assess the value of an adapted policy alongside the search information it provides. The distinction matters when different procedures achieve comparable best outcomes but differ in how frequently they generate useful candidates. Extending this analysis to several interacting design variables is a natural next step, including constrained multi-objective material design \cite{peng2023machine}. The two settings show how scientific feedback can revise inherited search preferences, increasing generation beyond a supervised convention while improving candidate quality.

\section{Methods}\label{sec:methods}

\subsection{Experimental design and evaluated distributions}

Let $t$ denote a task, $u$ a requested scientific target and $z$ an evaluated candidate. We define a feature map $\phi_t(z)$ and the set $S_t$ of its values represented in the supervised corpus. For reactor lattices, $\phi(z)$ is absorber count and $S=\{16\}$. For chemistry, $\phi_t(z)$ is ligand identity and $S_t=\{a(t)\}$, where $a(t)$ is a task-specific anchor ligand. The relevant probability is
\begin{equation}
 M_\theta(t,u)=\Pr_{z\sim q_\theta(\cdot\mid t,u)}[\phi_t(z)\notin S_t],
 \label{eq:mass}
\end{equation}
where $q_\theta$ is the distribution of candidates passed to the evaluator. The set $S_t$ records the feature values represented in the supervised corpus. Equation~\ref{eq:mass} therefore measures probability assigned to choices outside the declared training convention. Those choices are available in the decoding space from initialization.

We measure candidate quality alongside off-corpus probability. Exact enumeration gives expected absolute target error in chemistry. Unselected corrected samples give target-window incidence in the lattice experiment. Best-of-budget objectives measure the selected outcome of each search. Both online update rules receive 2{,}000 evaluations per run. Table~\ref{tab:settings} summarizes the two settings.

\begin{table}[htbp]
\centering
\caption{Two experimental settings and their measurements. Each supervised corpus fixes an identifiable feature, while online generation permits alternative values.}
\label{tab:settings}
\small
\setlength{\tabcolsep}{4pt}
\begin{tabular}{>{\raggedright\arraybackslash}p{0.20\textwidth}>{\raggedright\arraybackslash}p{0.34\textwidth}>{\raggedright\arraybackslash}p{0.34\textwidth}}
\toprule
 & Measured-reaction catalog & Simulator-evaluated lattices\\
\midrule
Policy & Masked categorical autoregressive network, hidden size 48 & Gemma 3 sequence backbone, 270 million parameters\\
Supervised convention & One anchor ligand per aryl-halide task & Exactly 16 absorber rods per lattice\\
Online candidate space & All 3{,}955 measured catalog rows, conditioned on task & Fuel/absorber choices at 264 non-guide positions\\
Feedback & Deterministic lookup of measured yield & Stochastic continuous-energy Monte Carlo transport\\
Distribution measurement & Exact task-conditioned catalog probabilities & Corrected sample incidence and selected-design trajectories\\
Quality measurement & Expected target error and catalog optima & Target-window incidence and best-of-budget objective\\
Replication & Five seeds, two initialization histories and two target schedules & Five seeds for principal searches, controls and the initialization ablation\\
\bottomrule
\end{tabular}
\end{table}

In chemistry, a trie mask limits candidates to measured reactions. In the reactor experiment, $y$ denotes raw generated text and $C(y)$ its deterministic correction. The evaluated distribution is the pushforward
\begin{equation}
 q_\theta(z\mid t,u)=\sum_{y:C(y)=z}p^{\mathrm{dec}}_\theta(y\mid t,u).
 \label{eq:pushforward}
\end{equation}
Here $p^{\mathrm{dec}}_\theta$ is the raw-text decoding distribution with its sampling filters, stopping rules and generation cap. The model likelihood used for updates is $\pi_\theta$. We use online variants of DPO \cite{rafailov2023direct} and GRPO \cite{shao2024deepseekmath,williams1992simple} to test pairwise and group-relative adaptation under the same scientific feedback. Reactor evaluation samples from the corrected distribution $q_\theta$, and the implemented updates use the sequence likelihood $\log\pi_\theta(C(y)\mid t,u)$.

\subsection{Measured-reaction experiment}\label{sec:methods_empirical}

We use the published Buchwald--Hartwig measurements \cite{ahneman2018predicting} distributed in the \texttt{Plates1-3} worksheet of the upstream \texttt{rxn\_yields} repository. This subset contains 3{,}955 rows spanning 15 aryl-halide tasks, four ligands, 22 additives and three bases. Five missing cells in its nominal factorial grid are excluded without imputation. We sort task and ligand identifiers and assign the anchor ligand by $t\bmod4$, independently of yield. The restricted catalog contains 990 measured rows, approximately 66 per task.

The categorical autoregressive policy has hidden size 48. It sequentially generates ligand, additive and base, conditioned on the task and either a requested yield or an unconditional pre-training token. A prefix trie masks each softmax to measured rows for that task. This small non-language architecture permits exact enumeration of candidate probabilities.

For the categorical network, the stage labeled CPT is unconditional pre-training from random weights. It runs for five epochs by maximum likelihood on the restricted rows. SFT runs for ten epochs using the eight restricted rows nearest to each requested yield per task. Each target schedule therefore defines its own SFT examples as well as its online rewards. Both CPT$+$SFT and SFT-only arms are repeated for five seeds. AdamW learning rates are $10^{-3}$ for supervised stages and $3\times10^{-4}$ for online updates. Single-target alignment requests yield 90. Multi-target alignment cycles through a random permutation of ten requested yields, 10,20,$\ldots$,100. Reward is
\begin{equation}
 r(z;u)=-|y_{\mathrm{measured}}(z)-u|/100.
\end{equation}
DPO uses 1{,}000 steps with two samples per step and $\beta=0.01$. GRPO uses 500 steps with four samples per step. Both use temperature 1.0 and 2{,}000 reward lookups. The budget counts each lookup, including repeated evaluations of the same measured row.

For each task and target, we enumerate all measured candidates to calculate exact expected absolute error and off-anchor mass. Aggregate values average the specified task and target schedule. Seed uncertainty describes variation among independently trained policies. The full and restricted catalog bounds use the lowest absolute measured error for each task and target in the corresponding catalog. Changes are paired to the same-seed SFT initialization. Supplementary Information reports both initialization arms and data provenance.

\subsection{Computational context of the lattice testbed}

Reactor studies provide established examples of optimization over discrete scientific configurations. Previous work has examined soluble-boron-free lattice and core designs with centrally shielded burnable absorbers \cite{nguyen2021truly,nguyen2019advanced,wijaya2024multi}. Particle swarm optimization has been applied to seed-blanket reactor configurations \cite{stefani2026optimization}, and hybrid adjoint and genetic methods have optimized core loading patterns \cite{abuzlf2025merging}. Genetic algorithms have also addressed load-following control in multi-unit SMR systems \cite{byun2025genetic}. These applications provide the computational setting for the lattice experiment.

Learning-based approaches span assembly design through physics-informed reinforcement learning \cite{radaideh2021physics} and AI-based core design \cite{sobes2021ai}. Convolutional networks have supported fuel-reloading optimization in PWRs and block-type high-temperature gas-cooled reactors \cite{wan2022optimization,li2023development}. Combining core physics with machine learning has improved parameter prediction in boiling-water-reactor operations \cite{oktavian2024integrating}. Neural surrogates have supported optimization of crud deposition \cite{andersen2022novel} and coupled fast/thermal experiments \cite{pevey2022neural}. Combined neural-network and genetic approaches have also optimized assemblies using neutronic and thermal-hydraulic quantities \cite{rishehri2023design}. This range of applications motivates a testbed in which a scientific evaluator can directly guide learning.

Language-model applications in this domain include transient identification \cite{qi2024multimodal}, shutdown-initiating-event classification \cite{xian2025knowledge} and reactor-operation assistance \cite{lee2025large}. Language models have also supported explainable fault diagnosis in complex systems \cite{dave2024integrating}. Here, the sequence policy generates complete configurations, and scientific evaluations supply each online update. Together with the categorical chemistry policy, this setup tests the effect of feedback on the generative distribution across two model architectures.

\subsection{Stochastic lattice evaluator and objective}

The domain is a $17\times17$ PWR-type assembly representative of lattices used in SMR concepts \cite{IAEA2024SMR}. Twenty-five guide and instrument positions are water-filled channels under all-rods-out conditions. The remaining 264 positions contain ordinary fuel or Gd-bearing fuel. Each source-corpus lattice contains 16 Gd-bearing rods and 248 ordinary fuel rods. Continuous-energy OpenMC explicitly resolves the pellet, gap, cladding and moderator using ENDF/B-VII.1 cross sections \cite{romano2015openmc,chadwick2011endf}. Reflective boundaries give $k_\infty$, stored under the column label $k_{\mathrm{eff}}$. Supplementary Information provides geometry and materials.

Corpus generation, alignment and matched searches use 20{,}000 particles per batch with ten inactive and 20 active batches, giving 400{,}000 active histories. The corpus mean reported standard deviation is 150.5 pcm and the median is 149 pcm. Pin fission rates are scored on the lattice mesh. For mesh powers $P_{ij}$, the implemented proxies are $F_q=\max_{ij}P_{ij}/\bar P$ and $F_{\Delta H}=\max_j\bar P_j/\bar P$. Here $\bar P_j$ averages column $j$, and zero-fission guide cells are included in the averages.

Equation~\ref{eq:fitness} combines the power-peaking proxies with target error. The objective has no count, novelty or diversity term. The single-target prompt requests $k=1.05$ together with $F_q=1.00$ and $F_{\Delta H}=1.00$. All online rewards use this beginning-of-life model with fixed material conditions. Separate depletion calculations characterize the evolution of selected layouts in Supplementary Information.

\subsection{Sequence policy and initialization}

The 10{,}000-pattern source corpus is split 80/10/10 into training, validation and test subsets. CPT and SFT use the same split, and every subset retains the 16-absorber convention. Raster serialization uses \texttt{f}, \texttt{g} and \texttt{c} for cell symbols, with spaces between cells and a newline after each row. Each lattice contains 289 cell symbols. A complete prompt and response contains 358 tokens under the Gemma tokenizer, whose vocabulary size is 256{,}000. The maximum training sequence length is 384 tokenizer tokens.

The backbone is Gemma 3 270M \cite{kamath2025gemma}. CPT uses unconditional lattice sequences for five epochs. SFT then uses prompts containing evaluated $k$, $F_q$ and $F_{\Delta H}$ for ten epochs. This staged training establishes a lattice prior before conditioning \cite{bengio2009curriculum,gururangan2020don}. Both stages use full fine-tuning with AdamW \cite{loshchilov2017decoupled} at learning rate $5\times10^{-5}$. The per-device batch size is four with four accumulation steps, giving an effective batch of 16. Training uses gradient checkpointing and bfloat16 weights. The SFT-only arm follows the same supervised setup with CPT omitted. Online updates use learning rate $10^{-5}$ and temperature 1.0.

The principal corpus, training, alignment and GA campaigns use an Intel Core i5-12400F CPU and NVIDIA RTX 3070 GPU. Supplemental objective-removal and random-search campaigns use an Apple M5 Max workstation, as do the generation audit and depletion calculations. All matched transport evaluations use the specified history counts. Each evaluation takes approximately a minute on the reported desktop setting.

\subsection{Implemented online update rules}\label{sec:updates}

The pairwise variant draws two candidates per prompt, evaluates both and orders them by the lower-is-better scientific objective. With winning and losing sequences $z_w,z_l$, its loss is
\begin{equation}
 \mathcal L_{\mathrm{pair}}=-\log\sigma\{\beta[\log\pi_\theta(z_w\mid t,u)-\log\pi_\theta(z_l\mid t,u)]\},\qquad \beta=0.01.
 \label{eq:pair}
\end{equation}
We use a reference-free online DPO variant \cite{rafailov2023direct} with the pairwise logistic loss above. One thousand two-candidate steps give 2{,}000 evaluations. The scientific evaluator supplies fresh candidate rankings at each step.

The group-relative variant draws $G=4$ candidates. If $\ell_i$ is the objective value, with in-group mean $\mu_G$ and standard deviation $\sigma_G$, the advantage and loss are
\begin{align}
 A_i&=(\mu_G-\ell_i)/(\sigma_G+10^{-8}),\\
 \mathcal L_{\mathrm{group}}&=-G^{-1}\sum_{i=1}^G A_i\log\pi_\theta(z_i\mid t,u).
 \label{eq:group}
\end{align}
Objective values below the group mean give positive advantages. This GRPO variant \cite{shao2024deepseekmath} uses a group-relative REINFORCE-style loss \cite{williams1992simple}. The update directly weights sequence log likelihoods by these group-relative advantages. Gradients are clipped to a global norm of 1.0. Five hundred four-candidate steps give 2{,}000 evaluations. DPO and GRPO throughout the manuscript denote these implemented variants.

Raw reactor text is filtered to the lattice alphabet and padded with \texttt{f} or truncated to 289 cell symbols. The 25 guide coordinates are overwritten with \texttt{c}, and stray \texttt{c} symbols are replaced with \texttt{f}. The resulting sequence is used for both transport and the likelihood in Equations~\ref{eq:pair} and \ref{eq:group}. Every \texttt{g} symbol comes from generation.

\subsection{Random-search, evolutionary and objective-removal controls}

Uninformed random search draws count uniformly from integers 0--264 and chooses that many non-guide positions uniformly without replacement. Informed random search uses the same placement rule and draws count from 20--40. Its interval is selected after observing alignment, so this control explicitly supplies the information identified in the study. Each procedure has five runs of 2{,}000 evaluations. All samples contribute to target-window counts. Best-of-budget comparisons use the best objective in each run.

Both GAs use a population of 40 with tournament selection of size three \cite{goldberg1989genetic}. They apply position-wise crossover with probability 0.5 and retain two elites. Fifty evaluated populations give 2{,}000 recorded evaluations. Repair and swap mutation preserve 16 absorbers in the constrained GA. The unconstrained GA initializes count uniformly from 10--30 and applies bit-flip mutation with probability 0.05 per position. Fixed guide coordinates are restored after crossover. Both GAs use the online objective and simulator specified for alignment. We also construct symmetric 16- and 24-absorber reference layouts for comparison.

The objective-removal experiment sets the target-penalty weight to zero in five matched single-target DPO runs. Each uses 1{,}000 updates with two evaluations per update and the same-seed CPT$+$SFT initialization. The prompt, learning rate, temperature, correction and transport setup are held fixed. We compare selected-design inventories averaged over the final 100 steps against the corresponding full-objective statistic. The exceptional high-inventory seed remains included in the mean and dispersion.

\subsection{Unselected generation and high-statistics checks}

Raw decoding uses temperature 1.0 with top-$k$ 64 and nucleus probability 0.95 inherited from the Gemma 3 generation configuration, a 400-token generation cap and the tokenizer's end-of-sequence token. The subsequent correction step sets the lattice length to 289 cell symbols. The raw exact-length validity rate is 0.0\%. Parser-compatible yield is 65.3\%, and 92.3\% of outputs require guide-coordinate replacement after length normalization. These measurements characterize the text-generation and correction stages separately. Of 10{,}000 raw outputs, 8{,}878 exceed 289 lattice symbols and 1{,}111 fall short. The 11 outputs with exactly 289 lattice symbols also contain non-lattice characters.

The generation audit contains 1{,}000 outputs from each of ten final single-target CPT$+$SFT checkpoints at temperature 1.0. Strict raw validity requires exactly 289 cell symbols after ignoring formatting spaces and newlines, with no remaining non-lattice characters. This criterion is applied before filtering or correction. Parser compatibility permits at least 289 valid symbols and deterministic retention of the first 289, again with no remaining non-lattice characters after formatting is ignored. Guide-coordinate replacements are recorded after length normalization. We randomly sample 50 corrected outputs per checkpoint for transport evaluation, without fitness ranking. For the supervised reference in Fig.~\ref{fig:concept}c, the same generation and correction procedure draws 1{,}000 outputs from each seed-specific CPT$+$SFT checkpoint before alignment and records corrected inventory without transport evaluation. The aligned 101/500, inventory-uniform 36/10{,}000 and informed-random 521/10{,}000 target-window rates use the same online transport settings. Their ratios compare the reported sampling procedures.

Each best layout from the five DPO and five GRPO runs receives an independent evaluation with 200{,}000 particles per batch, ten inactive batches and 100 active batches. The 20 million active histories provide higher-statistics estimates for each fixed selected layout. Supplementary Information reports all per-layout results.

\subsection{Statistical units and computational accounting}

Principal searches and initialization ablations report means and sample standard deviations across five training seeds. Exact two-sided Mann--Whitney tests compare best-of-budget values. We interpret nonsignificant comparisons together with the observed performance values. For five-run trajectory bands, runs are interpolated to 400 equally spaced evaluation counts over common support. Pointwise intervals are $\bar x\pm t_{0.975,4}s/\sqrt5$. Chemistry changes are paired to the same-seed initialization, and per-policy expectations use exact catalog enumeration. Reactor target-window rates use the sample counts specified above.

Each online search uses 2{,}000 evaluator calls. Chemistry counts repeated measured-yield lookups, while the reactor uses newly executed transport evaluations. Reactor policies also use a shared corpus of 10{,}000 physics evaluations and supervised training. Corpus construction, training, high-statistics reevaluation, generation audits and depletion calculations are recorded separately from the matched online search budget.

\backmatter

\bmhead{Acknowledgements}
We thank the maintainers of the open-source OpenMC Monte Carlo neutron transport code and the Gemma~3 language model for making their tools publicly available.

The authors used large language model tools to assist with manuscript organization, consistency checks and language editing. The authors reviewed and edited the resulting text and take full responsibility for the publication.

\section*{Declarations}

\bmhead{Funding}
This work was supported by the U.S. Department of Energy.

\bmhead{Competing interests}
The authors declare no competing interests.

\bmhead{Ethics approval and consent to participate}
This study uses computational simulations and a published measured-reaction dataset and does not involve human participants, human tissue, or live animals. Therefore, ethics approval is not applicable.

\bmhead{Consent for publication}
Not applicable.

\bmhead{Data availability}
The $10^4$ OpenMC-evaluated training configurations are available in the Hugging Face repository \url{https://huggingface.co/datasets/Sixticket/ReactorGen-Dataset}, in the file \texttt{reactor\_10k\_final.csv}. Source data and scripts for the result figures accompany the submission. The Buchwald--Hartwig table is the published dataset of Ahneman et al.\ \cite{ahneman2018predicting}. We use the \texttt{Plates1-3} worksheet distributed in the \texttt{rxn\_yields} repository without imputation or deduplication. Its SHA-256 checksum is recorded in the code repository. The table remains available upstream under its original terms.

\bmhead{Code availability}
The implementation is available at \url{https://github.com/sixticket/reactor_design_optimization}. It includes OpenMC model construction and dataset generation, the CPT, SFT, DPO and GRPO pipelines, genetic-algorithm and random-search controls, objective-removal and generation-yield analyses, and the measured-yield chemistry experiment.

\bmhead{Author contributions}
Y.P.L. conceived the study and developed the methodology. Y.P.L. performed the simulations and model development, analyzed the results and wrote the original manuscript. S.R. and S.C. contributed to interpretation and reviewed and edited the manuscript. S.B.A. supervised the research and contributed to its conceptualization. S.B.A. also reviewed and edited the manuscript. All authors read and approved the final manuscript.


\bibliography{sn-bibliography}

@article{chen2026xray,
  author  = {Chen, Zhantao and Petsch, Alexander N. and Israelski, Aidan J. and
             Plumley, Rajan and Shen, Lingjia and Wang, Cong and Peng, Cheng and
             Ni, Yuan and Bansil, Arun and Chowdhury, Sugata and Li, Mingda and
             Thayer, Jana B. and Thampy, Vivek and Turner, Joshua J.},
  title   = {An agentic artificially intelligent X-ray scientist},
  journal = {Nature Machine Intelligence},
  volume  = {8},
  pages   = {1075--1086},
  year    = {2026},
  doi     = {10.1038/s42256-026-01261-5}
}

@article{shi2026matbrain,
  author  = {Shi, Tongyu and Li, Yutang and Li, Zhanyuan and Liu, Qian and
             Zhou, Jie and Xu, Wenhe and Li, Yang and Dai, Dawei and He, Rui and
             Zhou, Wenhua and Wang, Jiahong and Yu, Xue-Feng},
  title   = {A collaborative agent with two lightweight synergistic models for
             autonomous crystal materials research},
  journal = {Nature Machine Intelligence},
  year    = {2026},
  doi     = {10.1038/s42256-026-01298-6}
}

@article{nmi2026transparency,
  title   = {Multi-agent {AI} systems need transparency},
  journal = {Nature Machine Intelligence},
  volume  = {8},
  pages   = {1},
  year    = {2026},
  doi     = {10.1038/s42256-026-01183-2}
}

@article{romano2015openmc,
  title={OpenMC: A state-of-the-art Monte Carlo code for research and development},
  author={Romano, Paul K and Horelik, Nicholas E and Herman, Bryan R and Nelson, Adam G and Forget, Benoit and Smith, Kord},
  journal={Annals of Nuclear Energy},
  volume={82},
  pages={90--97},
  year={2015},
  publisher={Elsevier}
}

@article{zheng2023unifying,
  title={Unifying the design space and optimizing linear and nonlinear truss metamaterials by generative modeling},
  author={Zheng, Li and Karapiperis, Konstantinos and Kumar, Siddhant and Kochmann, Dennis M},
  journal={Nature Communications},
  volume={14},
  number={1},
  pages={7563},
  year={2023},
  publisher={Nature Publishing Group UK London}
}

@book{IAEA2024SMR,
  author    = "{International Atomic Energy Agency}",
  title     = "Small {M}odular {R}eactor {T}echnology {C}atalogue 2024 {E}dition",
  publisher = "IAEA",
  address   = "Vienna",
  year      = "2024",
  note      = "A Supplement to: Small Modular Reactors: Advances in Developments 2024"
}

@article{rishehri2023design,
  title={Design and optimization of dual-cooled fuel assembly in a 12$\times$ 12 configuration for NuScale SMR based on neutronic-thermohydraulic parameters using the combined ANN-GA approach},
  author={Rishehri, Hossein Zayermohammadi and others},
  journal={Progress in Nuclear Energy},
  volume={163},
  pages={104799},
  year={2023},
  publisher={Elsevier}
}

@article{stefani2026optimization,
  title={Optimization of a NuScale-like SMR: A Master-Slave Parallel computing in Particle Swarm approach for enhancing nuclear reactor project and initial fuel cycle efficiency in seed-blanket configurations},
  author={Stefani, Giovanni L and da Silva, Marcelo V and Gon{\c{c}}alves, Diego Manoel E and Guimar{\~a}es, Jo{\~a}o Victor SA and de Souza, Diego JSN and Caixeta, Bernardo M and Santos, Marcelo C and de Lima, Alan MM and Nicolau, Andressa S and Pereira, Claudio MNA and others},
  journal={Progress in Nuclear Energy},
  volume={192},
  pages={106119},
  year={2026},
  publisher={Elsevier}
}

@article{nguyen2021truly,
  title={Truly-optimized PWR lattice for innovative soluble-boron-free small modular reactor},
  author={Nguyen, Xuan Ha and Jang, Seongdong and Kim, Yonghee},
  journal={Scientific reports},
  volume={11},
  number={1},
  pages={12891},
  year={2021},
  publisher={Nature Publishing Group UK London}
}

@article{nguyen2019advanced,
  title={An advanced core design for a soluble-boron-free small modular reactor ATOM with centrally-shielded burnable absorber},
  author={Nguyen, Xuan Ha and Kim, ChiHyung and Kim, Yonghee},
  journal={Nuclear Engineering and Technology},
  volume={51},
  number={2},
  pages={369--376},
  year={2019},
  publisher={Elsevier}
}

@article{wijaya2024multi,
  title={Multi-batch core design study for innovative small modular reactor based on centrally-shielded burnable absorber},
  author={Wijaya, Steven and Nguyen, Xuan Ha and Jeong, Yunseok and Kim, Yonghee},
  journal={Nuclear Engineering and Technology},
  volume={56},
  number={3},
  pages={907--915},
  year={2024},
  publisher={Elsevier}
}

@article{abuzlf2025merging,
  title={Merging adjoint-based determinism with genetic algorithms: A hybrid approach to reactor core loading pattern optimization},
  author={Abuzlf, Hadi and Gilad, Erez},
  journal={Annals of Nuclear Energy},
  volume={224},
  pages={111664},
  year={2025},
  publisher={Elsevier}
}

@article{byun2025genetic,
  title={Genetic algorithm-based non-synchronous unit-specific optimal load-following control of multi-unit small modular reactor},
  author={Byun, Hyeon-Ho and Yim, Man-Sung},
  journal={Energy},
  volume={314},
  pages={134091},
  year={2025},
  publisher={Elsevier}
}

@article{radaideh2021physics,
  title={Physics-informed reinforcement learning optimization of nuclear assembly design},
  author={Radaideh, Majdi I and Wolverton, Isaac and Joseph, Joshua and Tusar, James J and Otgonbaatar, Uuganbayar and Roy, Nicholas and Forget, Benoit and Shirvan, Koroush},
  journal={Nuclear Engineering and Design},
  volume={372},
  pages={110966},
  year={2021},
  publisher={Elsevier}
}

@article{sobes2021ai,
  title={AI-based design of a nuclear reactor core},
  author={Sobes, Vladimir and Hiscox, Briana and Popov, Emilian and Archibald, Rick and Hauck, Cory and Betzler, Ben and Terrani, Kurt},
  journal={Scientific reports},
  volume={11},
  number={1},
  pages={19646},
  year={2021},
  publisher={Nature Publishing Group UK London}
}

@article{andersen2022novel,
  title={A novel method for controlling crud deposition in nuclear reactors using optimization algorithms and deep neural network based surrogate models},
  author={Andersen, Brian and Hou, Jason and Godfrey, Andrew and Kropaczek, Dave},
  journal={Eng},
  volume={3},
  number={4},
  pages={504--522},
  year={2022},
  publisher={MDPI}
}

@article{wan2022optimization,
  title={Optimization method of fuel-reloading pattern for PWR based on the improved convolutional neural network and genetic algorithm},
  author={Wan, Chenghui and Lei, Kaihui and Li, Yisong},
  journal={Annals of Nuclear Energy},
  volume={171},
  pages={109028},
  year={2022},
  publisher={Elsevier}
}

@article{li2023development,
  title={Development and research of triangle-filter convolution neural network for fuel reloading optimization of block-type HTGRs},
  author={Li, Zhan and Wang, Jincheng and Huang, Jie and Ding, Ming},
  journal={Applied Soft Computing},
  volume={136},
  pages={110126},
  year={2023},
  publisher={Elsevier}
}

@article{pevey2022neural,
  title={Neural Network Acceleration of Genetic Algorithms for the Optimization of a Coupled Fast/Thermal Nuclear Experiment},
  author={Pevey, John and Sobes, Vlad and Hines, Wes J},
  journal={Frontiers in Energy Research},
  volume={10},
  pages={874194},
  year={2022},
  publisher={Frontiers Media SA}
}

@article{oktavian2024integrating,
  title={Integrating core physics and machine learning for improved parameter prediction in boiling water reactor operations},
  author={Oktavian, M Rizki and Nistor, Jonathan and Gruenwald, J Thomas and Xu, Yunlin},
  journal={Scientific Reports},
  volume={14},
  number={1},
  pages={5835},
  year={2024},
  publisher={Nature Publishing Group UK London}
}

@article{elton2019deep,
  title={Deep learning for molecular design—a review of the state of the art},
  author={Elton, Daniel C and Boukouvalas, Zois and Fuge, Mark D and Chung, Peter W},
  journal={Molecular Systems Design \& Engineering},
  volume={4},
  number={4},
  pages={828--849},
  year={2019},
  publisher={Royal Society of Chemistry}
}

@article{bilodeau2022generative,
  title={Generative models for molecular discovery: Recent advances and challenges},
  author={Bilodeau, Camille and Jin, Wengong and Jaakkola, Tommi and Barzilay, Regina and Jensen, Klavs F},
  journal={Wiley Interdisciplinary Reviews: Computational Molecular Science},
  volume={12},
  number={5},
  pages={e1608},
  year={2022},
  publisher={Wiley Online Library}
}

@inproceedings{bengio2009curriculum,
  title     = {Curriculum Learning},
  author    = {Bengio, Yoshua and Louradour, J{\'e}r{\^o}me and Collobert, Ronan and Weston, Jason},
  booktitle = {Proceedings of the 26th Annual International Conference on Machine Learning},
  editor    = {Danyluk, Andrea Pohoreckyj and Bottou, L{\'e}on and Littman, Michael L.},
  pages     = {41--48},
  year      = {2009},
  publisher = {ACM},
  address   = {New York, NY, USA},
  doi       = {10.1145/1553374.1553380}
}

@article{gururangan2020don,
  title={Don't stop pretraining: Adapt language models to domains and tasks},
  author={Gururangan, Suchin and Marasovi{\'c}, Ana and Swayamdipta, Swabha and Lo, Kyle and Beltagy, Iz and Downey, Doug and Smith, Noah A},
  journal={arXiv preprint arXiv:2004.10964},
  year={2020}
}

@article{ouyang2022training,
  title={Training language models to follow instructions with human feedback},
  author={Ouyang, Long and Wu, Jeffrey and Jiang, Xu and Almeida, Diogo and Wainwright, Carroll and Mishkin, Pamela and Zhang, Chong and Agarwal, Sandhini and Slama, Katarina and Ray, Alex and others},
  journal={Advances in neural information processing systems},
  volume={35},
  pages={27730--27744},
  year={2022}
}

@article{wei2021finetuned,
  title={Finetuned language models are zero-shot learners},
  author={Wei, Jason and Bosma, Maarten and Zhao, Vincent Y and Guu, Kelvin and Yu, Adams Wei and Lester, Brian and Du, Nan and Dai, Andrew M and Le, Quoc V},
  journal={arXiv preprint arXiv:2109.01652},
  year={2021}
}

@article{rafailov2023direct,
  title={Direct preference optimization: Your language model is secretly a reward model},
  author={Rafailov, Rafael and Sharma, Archit and Mitchell, Eric and Manning, Christopher D and Ermon, Stefano and Finn, Chelsea},
  journal={Advances in neural information processing systems},
  volume={36},
  pages={53728--53741},
  year={2023}
}

@book{goldberg1989genetic,
  title={Genetic Algorithms in Search, Optimization, and Machine Learning},
  author={Goldberg, David E.},
  publisher={Addison-Wesley},
  address={Reading, MA},
  year={1989},
  isbn={9780201157673}
}

@article{kamath2025gemma,
  title={Gemma 3 technical report},
  author={Kamath, Aishwarya and Ferret, Johan and Pathak, Shreya and Vieillard, Nino and Merhej, Ramona and Perrin, Sarah and Matejovicova, Tatiana and Ram{\'e}, Alexandre and Rivi{\`e}re, Morgane and Rouillard, Louis and others},
  journal={CoRR},
  year={2025}
}

@article{vaswani2017attention,
  title={Attention is all you need},
  author={Vaswani, Ashish and Shazeer, Noam and Parmar, Niki and Uszkoreit, Jakob and Jones, Llion and Gomez, Aidan N and Kaiser, {\L}ukasz and Polosukhin, Illia},
  journal={Advances in neural information processing systems},
  volume={30},
  year={2017}
}

@article{qi2024multimodal,
  title={Multimodal learning using large language models to improve transient identification of nuclear power plants},
  author={Qi, Ben and Sun, Jun and Sui, Zhe and Xiao, Xingyu and Liang, Jingang},
  journal={Progress in Nuclear Energy},
  volume={177},
  pages={105421},
  year={2024},
  publisher={Elsevier}
}

@article{xian2025knowledge,
  title={A knowledge-informed large language model framework for US nuclear power plant shutdown initiating event classification for probabilistic risk assessment},
  author={Xian, Min and Wang, Tao and Zhang, Sai and Xu, Fei and Ma, Zhegang},
  journal={Proceedings of the Institution of Mechanical Engineers, Part O: Journal of Risk and Reliability},
  pages={1748006X251386900},
  year={2025},
  publisher={SAGE Publications Sage UK: London, England}
}

@article{lee2025large,
  title={Large language model agent for nuclear reactor operation assistance},
  author={Lee, Yoon Pyo and Cha, Joowon and Yu, Yonggyun and Kim, Seung Geun},
  journal={Nuclear Engineering and Technology},
  pages={103842},
  year={2025},
  publisher={Elsevier}
}

@article{dave2024integrating,
  title={Integrating LLMs for explainable fault diagnosis in complex systems},
  author={Dave, Akshay J and Nguyen, Tat Nghia and Vilim, Richard B},
  journal={arXiv preprint arXiv:2402.06695},
  year={2024}
}

@article{bastek2022inverting,
  title={Inverting the structure--property map of truss metamaterials by deep learning},
  author={Bastek, Jan-Hendrik and Kumar, Siddhant and Telgen, Bastian and Glaesener, Rapha{\"e}l N and Kochmann, Dennis M},
  journal={Proceedings of the National Academy of Sciences},
  volume={119},
  number={1},
  pages={e2111505119},
  year={2022},
  publisher={National Academy of Sciences}
}

@article{loshchilov2017decoupled,
  title={Decoupled weight decay regularization},
  author={Loshchilov, Ilya and Hutter, Frank},
  journal={arXiv preprint arXiv:1711.05101},
  year={2017}
}

@article{shao2024deepseekmath,
  title={Deepseekmath: Pushing the limits of mathematical reasoning in open language models},
  author={Shao, Zhihong and Wang, Peiyi and Zhu, Qihao and Xu, Runxin and Song, Junxiao and Bi, Xiao and Zhang, Haowei and Zhang, Mingchuan and Li, YK and Wu, Yang and others},
  journal={arXiv preprint arXiv:2402.03300},
  year={2024}
}

@article{chadwick2011endf,
  title={ENDF/B-VII. 1 nuclear data for science and technology: cross sections, covariances, fission product yields and decay data},
  author={Chadwick, Mark B and Herman, Michal and Oblo{\v{z}}insk{\`y}, P and Dunn, Michael E and Danon, Yaron and Kahler, AC and Smith, Donald L and Pritychenko, Boris and Arbanas, Goran and Arcilla, R and others},
  journal={Nuclear data sheets},
  volume={112},
  number={12},
  pages={2887--2996},
  year={2011},
  publisher={Elsevier}
}

@article{williams1992simple,
  title={Simple statistical gradient-following algorithms for connectionist reinforcement learning},
  author={Williams, Ronald J},
  journal={Machine learning},
  volume={8},
  number={3},
  pages={229--256},
  year={1992},
  publisher={Springer}
}

@article{guo2026sample,
  doi={10.1038/s42256-026-01200-4},
  title={Sample-efficient generative molecular design using memory manipulation},
  author={Guo, Jeff and Chen, Junwu and Gx-Chen, Anthony and Schwaller, Philippe},
  journal={Nature Machine Intelligence},
  volume={8},
  number={3},
  pages={449--460},
  year={2026},
  publisher={Nature Publishing Group UK London}
}

@article{zheng2026algebraic,
  title={Algebraic language models for inverse design of metamaterials via diffusion transformers},
  author={Zheng, Li and Kumar, Siddhant and Kochmann, Dennis M},
  journal={Nature Machine Intelligence},
  volume={8},
  number={4},
  pages={628--640},
  doi={10.1038/s42256-026-01218-8},
  year={2026},
  publisher={Nature Publishing Group UK London}
}

@article{park2026guiding,
  doi={10.1038/s42256-026-01262-4},
  title={Guiding generative models to uncover diverse and novel crystals via reinforcement learning},
  author={Park, Hyunsoo and Walsh, Aron},
  journal={Nature Machine Intelligence},
  volume={8},
  number={7},
  pages={1087--1099},
  year={2026},
  publisher={Nature Publishing Group UK London}
}

@article{antunes2024crystal,
  title={Crystal structure generation with autoregressive large language modeling},
  author={Antunes, Luis M and Butler, Keith T and Grau-Crespo, Ricardo},
  journal={Nature Communications},
  volume={15},
  number={1},
  pages={10570},
  year={2024},
  publisher={Nature Publishing Group UK London}
}

@article{xiao2023invertible,
  title={An invertible, invariant crystal representation for inverse design of solid-state materials using generative deep learning},
  author={Xiao, Hang and Li, Rong and Shi, Xiaoyang and Chen, Yan and Zhu, Liangliang and Chen, Xi and Wang, Lei},
  journal={Nature Communications},
  volume={14},
  number={1},
  pages={7027},
  year={2023},
  publisher={Nature Publishing Group UK London}
}

@article{arlt2026meta,
  title={Meta-designing quantum experiments with language models},
  author={Arlt, S{\"o}ren and Duan, Haonan and Li, Felix and Xie, Sang Michael and Wu, Yuhuai and Krenn, Mario},
  journal={Nature Machine Intelligence},
  volume={8},
  number={2},
  pages={148--157},
  year={2026},
  publisher={Nature Publishing Group UK London}
}

@article{maurizi2025designing,
  title={Designing metamaterials with programmable nonlinear responses and geometric constraints in graph space},
  author={Maurizi, Marco and Xu, Derek and Wang, Yu-Tong and Yao, Desheng and Hahn, David and Oudich, Mourad and Satpati, Anish and Bauchy, Mathieu and Wang, Wei and Sun, Yizhou and others},
  journal={Nature Machine Intelligence},
  volume={7},
  number={7},
  pages={1023--1036},
  year={2025},
  publisher={Nature Publishing Group UK London}
}

@article{mokaya2023testing,
  title={Testing the limits of SMILES-based de novo molecular generation with curriculum and deep reinforcement learning},
  author={Mokaya, Maranga and Imrie, Fergus and van Hoorn, Willem P and Kalisz, Aleksandra and Bradley, Anthony R and Deane, Charlotte M},
  journal={Nature Machine Intelligence},
  volume={5},
  number={4},
  pages={386--394},
  year={2023},
  publisher={Nature Publishing Group UK London}
}

@article{jia2019anthropogenic,
  title={Anthropogenic biases in chemical reaction data hinder exploratory inorganic synthesis},
  author={Jia, Xiwen and Lynch, Allyson and Huang, Yuheng and Danielson, Matthew and Lang’at, Immaculate and Milder, Alexander and Ruby, Aaron E and Wang, Hao and Friedler, Sorelle A and Norquist, Alexander J and others},
  journal={Nature},
  volume={573},
  number={7773},
  pages={251--255},
  year={2019},
  publisher={Nature Publishing Group UK London}
}

@article{peng2023machine,
  title={Machine learning-enabled constrained multi-objective design of architected materials},
  author={Peng, Bo and Wei, Ye and Qin, Yu and Dai, Jiabao and Li, Yue and Liu, Aobo and Tian, Yun and Han, Liuliu and Zheng, Yufeng and Wen, Peng},
  journal={Nature Communications},
  volume={14},
  number={1},
  pages={6630},
  year={2023},
  publisher={Nature Publishing Group UK London}
}

@article{ahneman2018predicting,
  doi={10.1126/science.aar5169},
  title={Predicting reaction performance in C--N cross-coupling using machine learning},
  author={Ahneman, Derek T and Estrada, Jes{\'u}s G and Lin, Shishi and Dreher, Spencer D and Doyle, Abigail G},
  journal={Science},
  volume={360},
  number={6385},
  pages={186--190},
  year={2018},
  publisher={American Association for the Advancement of Science}
}

@article{molesky2018inverse,
  title={Inverse design in nanophotonics},
  author={Molesky, Sean and Lin, Zin and Piggott, Alexander Y and Jin, Weiliang and Vuckovi{\'c}, Jelena and Rodriguez, Alejandro W},
  journal={Nature photonics},
  volume={12},
  number={11},
  pages={659--670},
  year={2018},
  publisher={Nature Publishing Group UK London}
}

@article{sanchez2018inverse,
  title={Inverse molecular design using machine learning: Generative models for matter engineering},
  author={Sanchez-Lengeling, Benjamin and Aspuru-Guzik, Al{\'a}n},
  journal={Science},
  volume={361},
  number={6400},
  pages={360--365},
  year={2018},
  publisher={American Association for the Advancement of Science}
}

@article{fawzi2022discovering,
  title={Discovering faster matrix multiplication algorithms with reinforcement learning},
  author={Fawzi, Alhussein and Balog, Matej and Huang, Aja and Hubert, Thomas and Romera-Paredes, Bernardino and Barekatain, Mohammadamin and Novikov, Alexander and R. Ruiz, Francisco J and Schrittwieser, Julian and Swirszcz, Grzegorz and others},
  journal={Nature},
  volume={610},
  number={7930},
  pages={47--53},
  year={2022},
  publisher={Nature Publishing Group UK London}
}

@article{geirhos2020shortcut,
  title={Shortcut learning in deep neural networks},
  author={Geirhos, Robert and Jacobsen, J{\"o}rn-Henrik and Michaelis, Claudio and Zemel, Richard and Brendel, Wieland and Bethge, Matthias and Wichmann, Felix A},
  journal={Nature Machine Intelligence},
  volume={2},
  number={11},
  pages={665--673},
  year={2020},
  publisher={Nature Publishing Group UK London}
}

@article{wang2023scientific,
  title={Scientific discovery in the age of artificial intelligence},
  author={Wang, Hanchen and Fu, Tianfan and Du, Yuanqi and Gao, Wenhao and Huang, Kexin and Liu, Ziming and Chandak, Payal and Liu, Shengchao and Van Katwyk, Peter and Deac, Andreea and others},
  journal={Nature},
  volume={620},
  number={7972},
  pages={47--60},
  year={2023},
  publisher={Nature Publishing Group UK London}
}

@article{sun2025accelerating,
  title={Accelerating protein engineering with fitness landscape modelling and reinforcement learning},
  author={Sun, Haoran and He, Liang and Deng, Pan and Liu, Guoqing and Zhao, Zhiyu and Jiang, Yuliang and Cao, Chuan and Ju, Fusong and Wu, Lijun and Liu, Haiguang and others},
  journal={Nature Machine Intelligence},
  volume={7},
  number={9},
  pages={1446--1460},
  year={2025},
  publisher={Nature Publishing Group UK London}
}

@article{rankovic2025large,
  title={Large language models as uncertainty-calibrated optimizers for experimental discovery},
  author={Rankovic, Bojana and Griffiths, Ryan-Rhys and Schwaller, Philippe},
  journal={Preprint at arXiv. https://doi. org/10.48550/arXiv},
  volume={2504},
  year={2025}
}

@article{torres2026generative,
  title={A generative artificial intelligence approach for peptide antibiotic optimization},
  author={Torres, Marcelo DT and Zeng, Yimeng and Wan, Fangping and Maus, Natalie and Gardner, Jacob and de la Fuente-Nunez, Cesar},
  journal={Nature Machine Intelligence},
  pages={1--16},
  year={2026},
  publisher={Nature Publishing Group UK London}
}

\end{document}


\title[Supplementary Information]{Supplementary Information for Scientific feedback reveals productive regimes beyond training conventions}
\author[1]{Yoon Pyo Lee}
\author[1]{Samrendra Roy}
\author[1,3,4]{Souvik Chakraborty}
\author[1,2]{Syed Bahauddin Alam\textsuperscript{*}}\email{alams@illinois.edu}
\affil[1]{\orgdiv{Department of Nuclear, Plasma, and Radiological Engineering}, \orgname{The Grainger College of Engineering, University of Illinois Urbana-Champaign}, \orgaddress{\city{Urbana}, \state{IL}, \country{USA}}}
\affil[2]{\orgdiv{National Center for Supercomputing Applications}, \orgname{University of Illinois Urbana-Champaign}, \orgaddress{\city{Urbana}, \state{IL}, \country{USA}}}
\affil[3]{\orgdiv{Department of Applied Mechanics}, \orgname{Indian Institute of Technology Delhi}, \orgaddress{\city{New Delhi}, \country{India}}}
\affil[4]{\orgdiv{Yardi School of Artificial Intelligence}, \orgname{Indian Institute of Technology Delhi}, \orgaddress{\city{New Delhi}, \country{India}}}
\maketitle
\renewcommand{\figurename}{Supplementary Fig.}
\renewcommand{\tablename}{Supplementary Table}

\section{Reactor implementation}
\label{si:physics}

Supplementary Table~\ref{si:parameters} gives the geometry, materials and transport settings supplementing main Methods. The 25 fixed channels comprise 24 control-rod thimbles and one central instrument tube. The 8~at\% Gd fraction relative to heavy-metal atoms is approximately 2.6~at\% relative to all atoms; the represented isotopes are $^{155}$Gd, $^{156}$Gd, $^{157}$Gd, $^{158}$Gd and $^{160}$Gd. Archived figure labels $k_{\mathrm{eff}}$ denote the reflective-lattice $k_\infty$.

Corpus transport seeds equal the pattern identifier plus one; additional campaigns use deterministic mappings recorded in their scripts. Inventory-constrained GA mutation swaps each Gd rod with ordinary fuel with probability 0.05.

\begin{table}[htbp]
\centering
\small
\caption{\textbf{Parameters of the two-dimensional reactor model.} The pin-resolved PWR-type assembly specifies the reflective-lattice, SMR-inspired testbed.}
\label{si:parameters}
\begin{tabular}{lll}
\toprule
Category & Parameter & Value\\
\midrule
\multirow{6}{*}{Geometry}
& Assembly & $17\times17$ PWR type\\
& Lattice pitch & 1.26~cm\\
& Fuel-pellet radius & 0.4096~cm\\
& Cladding inner diameter & 0.835~cm\\
& Cladding outer diameter & 0.950~cm\\
& Cladding material & Zircaloy-4\\
\midrule
\multirow{3}{*}{Corpus composition}
& Fixed guide/instrument channels & 25\\
& Gd-bearing fuel rods & 16\\
& Ordinary UO$_2$ rods & 248\\
\midrule
\multirow{6}{*}{Materials}
& UO$_2$ enrichment & 3.1~at\% $^{235}$U\\
& Ordinary UO$_2$ density & 10.298~g\,cm$^{-3}$\\
& Gd-bearing material density & 10.5~g\,cm$^{-3}$\\
& Gd fraction & 8~at\% of heavy-metal atoms\\
& Fuel--cladding gap & Void\\
& Moderator density & 0.701~g\,cm$^{-3}$\\
\midrule
\multirow{8}{*}{Transport}
& Energy treatment & Continuous energy\\
& Boundary condition & Reflective\\
& Thermal scattering & H in H$_2$O, $S(\alpha,\beta)$\\
& Initial fission source & Spatially uniform\\
& Particles per batch & 20,000\\
& Batches & 10 inactive $+$ 20 active\\
& Active histories & 400,000\\
& Nuclear data & ENDF/B-VII.1 \cite{chadwick2011endf}\\
\bottomrule
\end{tabular}
\end{table}

\section{Reactor controls and verification}
\label{si:optimization}
\label{si:inventory}

\subsection{Unselected generation and random sampling}

Of the 10,000 raw policy outputs audited in main Methods, 6,527 satisfy parser compatibility before guide-coordinate correction. Supplementary Table~\ref{si:incidence} gives corrected target-window incidence, including the inventory-conditioned comparison. Informed random sampling produces 112, 97, 92, 106 and 114 hits in its five runs; inventory-uniform sampling produces 6, 11, 10, 5 and 4. Eighteen aligned hits lie outside inventories 28--35.

\begin{table}[htbp]
\centering
\small
\caption{\textbf{Target-window incidence under the three sampling procedures.} Each entry gives hits/evaluated candidates and the corresponding percentage. The right column conditions both numerator and denominator on inventories 28--35. Policy candidates are corrected before transport evaluation. The pooled aligned row combines DPO and GRPO. Each random procedure has five runs of 2,000 evaluations, while each policy has five checkpoints with 50 randomly selected outputs per checkpoint.}
\label{si:incidence}
\begin{tabular}{lrr}
\toprule
Procedure & All inventories & Inventories 28--35\\
\midrule
Inventory-uniform random & 36/10,000 (0.36\%) & 36/281 (12.8\%)\\
Informed random & 521/10,000 (5.21\%) & 521/3,830 (13.6\%)\\
DPO & 33/250 (13.2\%) & 23/109 (21.1\%)\\
GRPO & 68/250 (27.2\%) & 60/173 (34.7\%)\\
Aligned pooled & 101/500 (20.2\%) & 83/282 (29.4\%)\\
\bottomrule
\end{tabular}
\end{table}

\subsection{Objective components and selected layouts}

\begin{figure}[H]
\centering
\includegraphics[width=\textwidth,height=0.78\textheight,keepaspectratio]{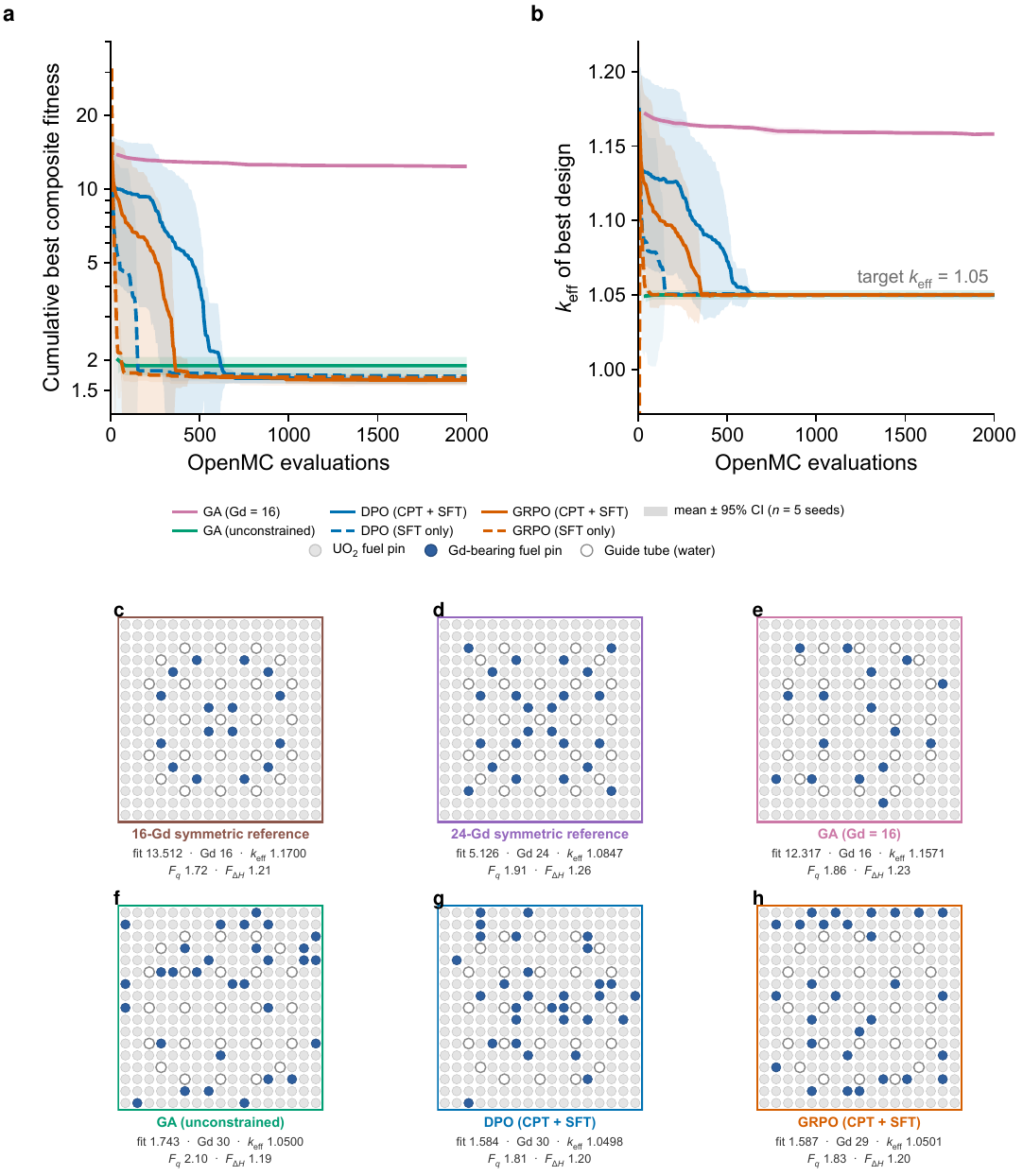}
\caption{\textbf{Additional convergence traces and selected lattice configurations.} \textbf{a,b} Cumulative-best composite fitness and the multiplication factor of the incumbent layout for five runs per method. Shading gives pointwise 95\% Student-$t$ intervals across seeds after interpolation. Solid lines denote CPT$+$SFT and dashed lines denote SFT only for the aligned policies. \textbf{c,d} Constructed symmetric references. \textbf{e--h} Best-of-budget layouts from the fixed-inventory GA, variable-inventory GA, DPO and GRPO, respectively. Values beneath each layout are its online evaluation. DPO/GRPO denote the implemented variants, and $k_{\mathrm{eff}}$ denotes $k_\infty$.}
\label{si:designs}
\end{figure}

For the best-of-budget designs in main Table~1, peaking-only fitness is $1.610\pm0.051$ for DPO, $1.616\pm0.037$ for GRPO and $1.714\pm0.043$ for variable-inventory GA. The GA target error is $181\pm138$~pcm. Fixed-inventory GA gives peaking-only fitness $1.572\pm0.043$ with approximately 10,800~pcm error; the symmetric 16-Gd reference gives 1.516 with approximately 12,000~pcm error. Informed random search gives $1.618\pm0.041$ with $35\pm13$~pcm error, and inventory-uniform search gives $1.712\pm0.025$ with $55\pm36$~pcm error. These components distinguish peaking performance from target attainment. Supplementary Fig.~\ref{si:designs} shows the additional convergence traces and selected layouts.

\subsection{Independent high-statistics verification}

Supplementary Table~\ref{si:high_statistics} reports all ten fixed-layout reevaluations with 20 million active histories per layout. Mean $F_q$ decreases from 1.882 to 1.554 for DPO and from 1.869 to 1.537 for GRPO. Eight of ten layouts improve in composite fitness.

\begin{table}[htbp]
\centering
\small
\setlength{\tabcolsep}{3pt}
\caption{\textbf{High-statistics reevaluation of the ten selected aligned layouts.} Each fixed layout receives one independent OpenMC evaluation with 20 million active histories. $\sigma_k$ is the OpenMC-reported Monte Carlo standard deviation in pcm, where one pcm is $10^{-5}$ in $k_\infty$. Composite fitness uses main-text Equation~(1). Seed identifiers refer to the five-seed campaign.}
\label{si:high_statistics}
\begin{tabular}{llrrrrrr}
\toprule
Method & Seed & Gd & $k_\infty$ & $\sigma_k$ (pcm) & $F_q$ & $F_{\Delta H}$ & $\mathcal L_{\mathrm{fit}}$\\
\midrule
DPO & 0 & 33 & 1.048970 & 22.0 & 1.564 & 1.234 & 1.535\\
DPO & 1 & 30 & 1.051477 & 19.0 & 1.549 & 1.225 & 1.567\\
DPO & 2 & 35 & 1.050032 & 20.1 & 1.629 & 1.145 & 1.439\\
DPO & 3 & 28 & 1.051875 & 19.3 & 1.544 & 1.151 & 1.574\\
DPO & 4 & 29 & 1.050691 & 19.6 & 1.483 & 1.252 & 1.460\\
GRPO & 0 & 28 & 1.052712 & 18.7 & 1.480 & 1.213 & 1.644\\
GRPO & 1 & 29 & 1.049501 & 22.1 & 1.520 & 1.161 & 1.427\\
GRPO & 2 & 29 & 1.052908 & 21.0 & 1.551 & 1.285 & 1.735\\
GRPO & 3 & 29 & 1.050547 & 20.2 & 1.593 & 1.239 & 1.507\\
GRPO & 4 & 29 & 1.050287 & 20.0 & 1.540 & 1.228 & 1.444\\
\bottomrule
\end{tabular}
\end{table}

\subsection{First passage and cumulative-best inventory reversals}

First passage is the first generated candidate with at least 20 Gd rods, the lower inventory boundary supplied to informed sampling. It measures inventory-threshold passage. Counts mark completion of the evaluation batch; the within-batch position is unresolved. DPO, GRPO and GA batches contain two, four and 40 candidates, respectively. Across five runs, DPO first-passage counts are $171.6\pm200.4$ and GRPO counts are $75.2\pm82.2$, or 8.6\% and 3.8\% of the online budget. Variable-inventory GA crosses in its first population in every run because its initial inventory range is 10--30.

\begin{figure}[H]
\centering
\includegraphics[width=\textwidth,height=0.78\textheight,keepaspectratio]{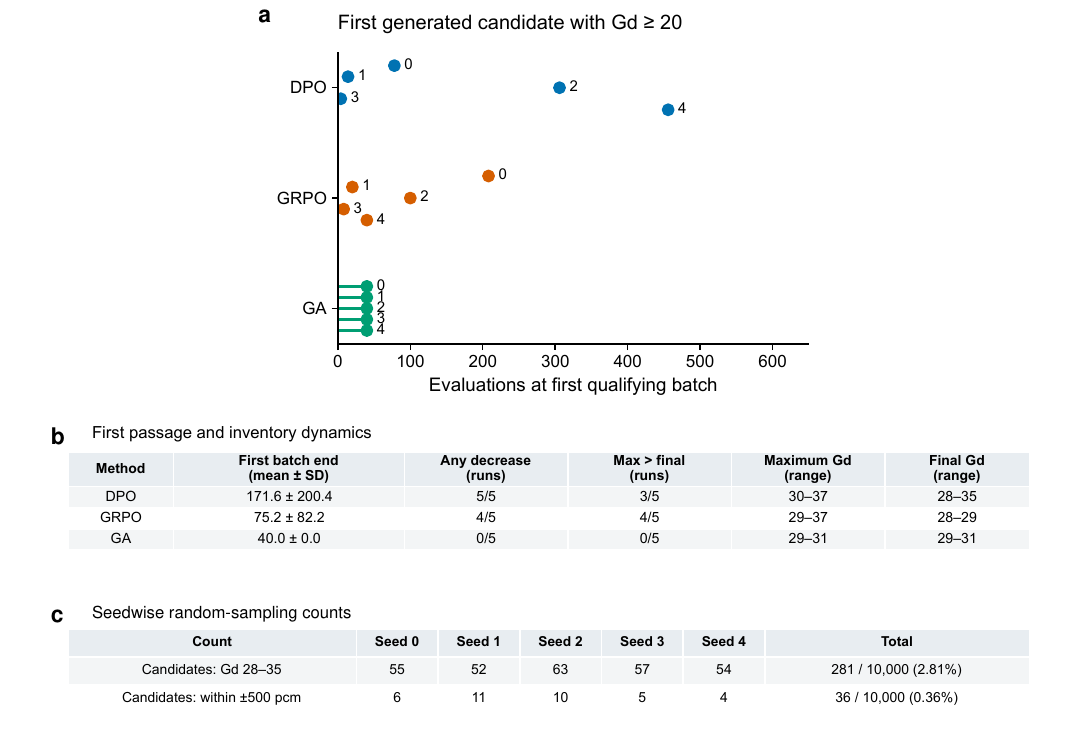}
\caption{\textbf{First passage, inventory reversals and seedwise sampling counts.} \textbf{a} First batch containing a candidate with at least 20 Gd rods. Points mark batch completion, segments show the unresolved within-batch interval, and numbers identify seeds. DPO, GRPO and GA batches contain 2, 4 and 40 candidates, respectively. \textbf{b} First-passage counts and subsequent changes in incumbent inventory across five runs. \textbf{c} Counts from five full-space random runs with 2,000 evaluations each. All 36 candidates within $|k_\infty-1.05|\leq0.005$ contain 28--35 Gd rods. DPO/GRPO denote the implemented variants.}
\label{si:inventory_figure}
\end{figure}

A reversal occurs when a strict recorded-fitness improvement replaces the incumbent with a lower-Gd layout. It occurs in all five DPO runs and four of five GRPO runs. Maximum incumbent inventories span 29--37 and final inventories 28--35; the maximum exceeds the final value in three DPO and four GRPO runs. DPO uses the logged running-best sequence. GRPO candidates are ordered by step and candidate index, retaining the earlier candidate on a recorded-fitness tie. Supplementary Fig.~\ref{si:inventory_figure} gives the per-seed summaries; main Figure~2a shows the incumbent trajectories.

\subsection{Peaking after target-penalty removal}

The matched five-seed intervention is reported in main Figure~3. The four zero-Gd runs reach peaking-only fitness approximately 1.5064. Across all five runs, including the higher-inventory exception, final-100-step chosen peaking fitness is $1.5574\pm0.1141$. These statistics use the preferred candidate within each generated pair.

\section{Initialization and prompt response}
\label{si:cpt}

\subsection{Five-seed initialization comparison}

The ablation compares SFT alone with CPT followed by SFT under the lattice training settings in main Methods. Supplementary Table~\ref{si:cpt_table} reports the complete best-design metrics. DPO best inventories are $31.2\pm1.9$ after SFT alone and $31.0\pm2.9$ after CPT$+$SFT; GRPO inventories are $36.0\pm8.4$ and $28.8\pm0.4$, respectively.

\begin{table}[htbp]
\centering
\small
\caption{\textbf{Best-design metrics for the reactor initialization ablation.} Mean $\pm$ sample standard deviation across five seeds. DPO and GRPO identify the implemented pairwise and group-relative variants.}
\label{si:cpt_table}
\begin{tabular}{llccc}
\toprule
Update & Initialization & Composite fitness & $F_q$ & $F_{\Delta H}$\\
\midrule
DPO & SFT only & $1.72\pm0.11$ & $1.95\pm0.08$ & $1.28\pm0.05$\\
DPO & CPT$+$SFT & $1.65\pm0.04$ & $1.88\pm0.11$ & $1.20\pm0.06$\\
GRPO & SFT only & $1.69\pm0.10$ & $1.98\pm0.19$ & $1.21\pm0.04$\\
GRPO & CPT$+$SFT & $1.66\pm0.05$ & $1.87\pm0.05$ & $1.24\pm0.03$\\
\bottomrule
\end{tabular}
\end{table}

\begin{figure}[H]
\centering
\includegraphics[width=\textwidth,height=0.64\textheight,keepaspectratio]{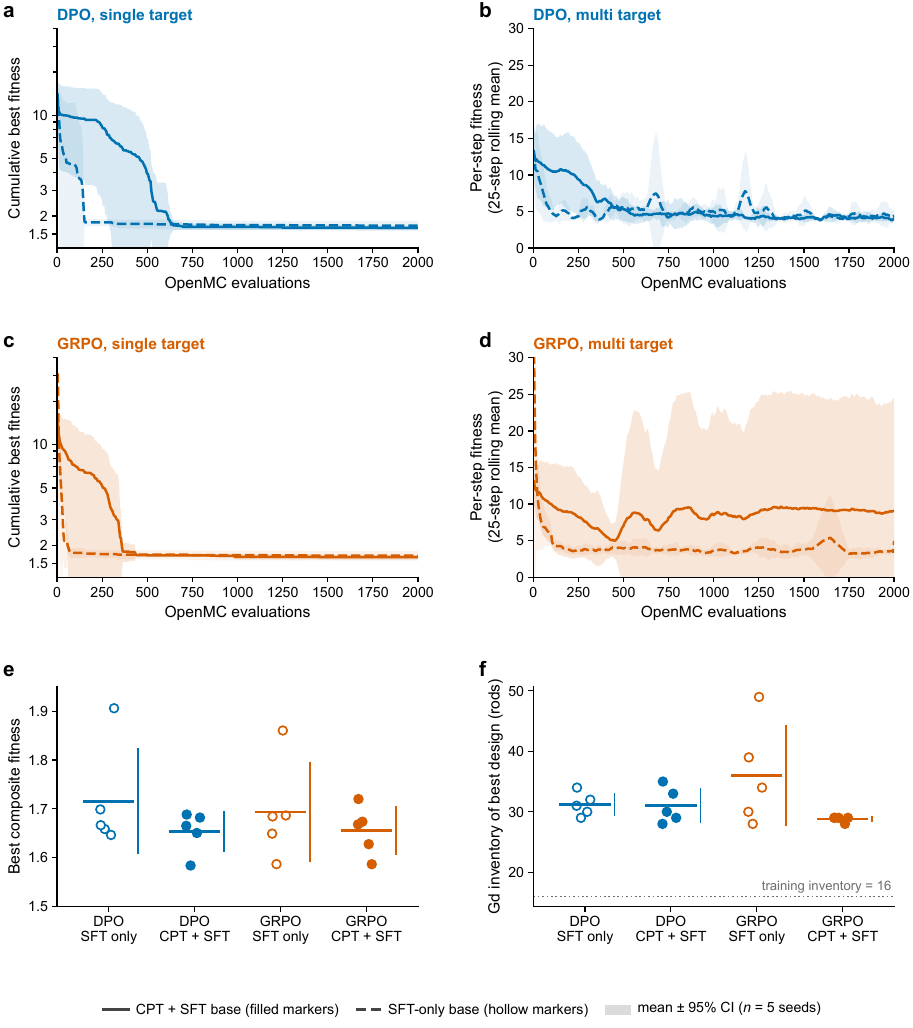}
\caption{\textbf{Initialization ablation.} Solid lines and filled markers denote CPT$+$SFT. Dashed lines and open markers denote SFT only. \textbf{a,c} Single-target cumulative-best fitness. \textbf{b,d} Multi-target per-step fitness with a 25-step rolling mean. \textbf{e,f} Best-design fitness and inventory by seed. Shading gives pointwise Student-$t$ intervals across five seeds. CPT reduces single-target fitness variation in these runs, with inventory and multi-target fitness effects that depend on the update variant. DPO/GRPO denote the implemented variants.}
\label{si:cpt_figure}
\end{figure}

Multi-target training requests $k\in[1.0,1.1]$. Each ten-step block draws one uniform target from each of ten equal bins and shuffles their order; peaking targets remain 1.00. DPO uses 1,000 steps with two candidates and GRPO 500 steps with four, giving 2,000 evaluations each. Supplementary Fig.~\ref{si:cpt_figure} shows single-target and multi-target trajectories for both initialization arms. Multi-target fitness is scored against each step's requested target.

\subsection{Archived prompt sweep}
\label{si:prompt}

The sweep uses one archived checkpoint per condition from runs preceding the five-seed campaign. These checkpoints are separate from the main-text seed0--4 models. The ten conditions comprise both initialization bases and their single-target and multi-target DPO/GRPO descendants; checkpoint identifiers are supplied in \texttt{data\_provenance.json}.

Each checkpoint generates 100 responses at each of seven requested $k$ values from 1.02 to 1.08 in increments of 0.01, below the supervised corpus range of 1.16192--1.21479. Sampling uses temperature 1.0, top-$k$ 64, top-$p$ 0.95, a 400-token limit and the tokenizer's end-of-sequence token. Parsing removes the prompt and non-lattice characters, retains the first 289 cell symbols and pads shorter responses with ordinary fuel. Gd counts are measured after length normalization and before guide-coordinate correction. Excluding responses with fewer than 50 valid symbols before padding retains 94--100 responses per target.

\begin{figure}[H]
\centering
\includegraphics[width=\textwidth,height=0.70\textheight,keepaspectratio]{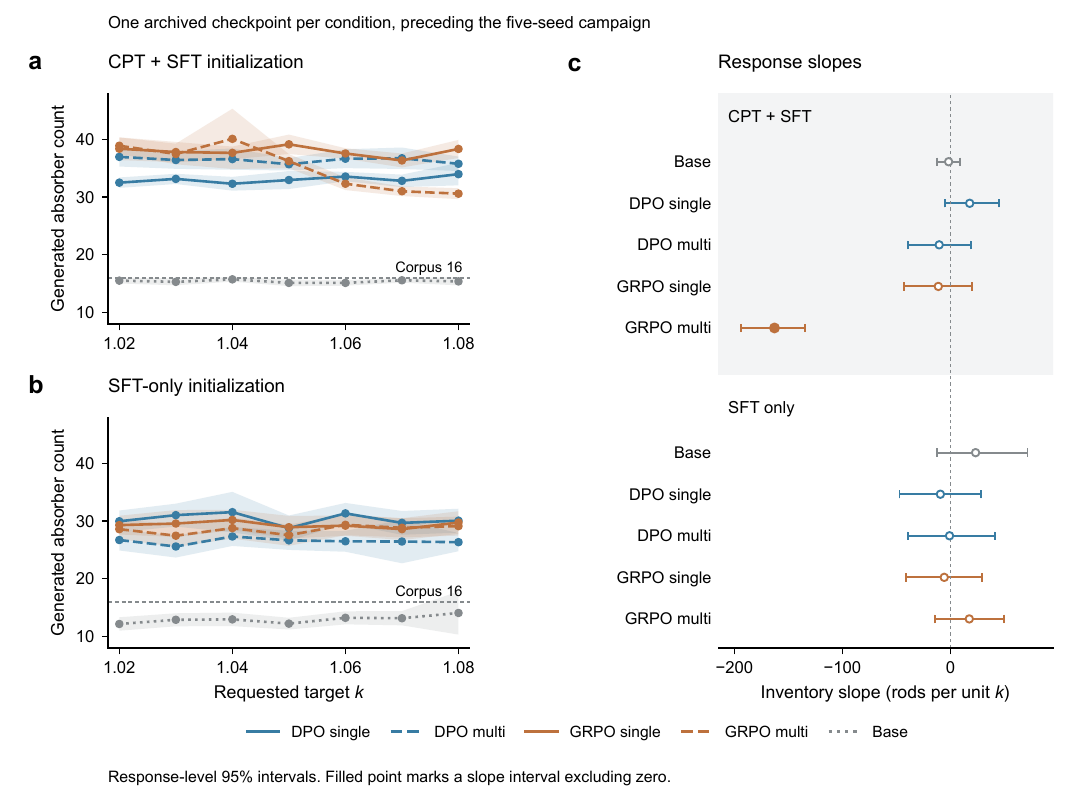}
\caption{\textbf{Prompt-conditioned inventory response in archived checkpoints.} \textbf{a,b} Mean generated absorber count for CPT$+$SFT and SFT-only checkpoints, respectively, at seven requested multiplication factors. Each condition uses one archived checkpoint from runs preceding the five-seed campaign. Each target begins with 100 responses; exclusion of responses containing fewer than 50 valid cell symbols before padding retains 94--100. Counts follow parsing and length normalization and precede guide-coordinate correction. Bands show two-sided 95\% Student-$t$ intervals across retained responses. \textbf{c} Slopes for all ten checkpoints with two-sided 95\% percentile intervals from 10,000 bootstrap resamples independently within each target cell. The filled point identifies the multi-target GRPO checkpoint with CPT$+$SFT whose interval excludes zero. Supplementary Table~\ref{si:prompt_slopes} includes the all-response sensitivity analysis.}
\label{si:steer_extra}
\end{figure}

We regress the seven targetwise mean Gd counts on requested $k$. A 95\% percentile interval uses 10,000 bootstrap resamples of responses independently within each target, preserving its retained count. These intervals describe response sampling at a fixed checkpoint. Multi-target GRPO with CPT$+$SFT has slope $-162.9$ rods per unit $k$, with interval $[-193.7,-134.3]$, corresponding to approximately ten fewer rods across the sweep. All other intervals overlap zero. Supplementary Table~\ref{si:prompt_slopes} includes the all-response sensitivity analysis, and Supplementary Fig.~\ref{si:steer_extra} shows the response curves.

\begin{table}[htbp]
\centering
\small
\setlength{\tabcolsep}{3pt}
\caption{\textbf{Prompt-response slopes and response-sampling intervals for all archived checkpoints.} Slopes are rods per unit requested multiplication factor. Each interval uses 10,000 bootstrap resamples within target cells at one fixed checkpoint. The filtered analysis excludes responses with fewer than 50 valid lattice symbols before padding. The all-response analysis retains all 700 responses per checkpoint.}
\label{si:prompt_slopes}
\begin{tabular}{lrrrr}
\toprule
& \multicolumn{2}{c}{Filtered responses} & \multicolumn{2}{c}{All responses}\\
Checkpoint condition & Slope & 95\% interval & Slope & 95\% interval\\
\midrule
CPT$+$SFT base & $-1.6$ & $[-12.2,9.1]$ & $0.4$ & $[-13.1,13.5]$\\
CPT$+$SFT / DPO single & $17.9$ & $[-5.1,45.1]$ & $22.5$ & $[-8.3,55.2]$\\
CPT$+$SFT / GRPO single & $-11.2$ & $[-42.6,20.1]$ & $-21.5$ & $[-56.9,12.9]$\\
CPT$+$SFT / DPO multi & $-10.4$ & $[-39.3,18.9]$ & $-14.4$ & $[-49.6,20.1]$\\
CPT$+$SFT / GRPO multi & $-162.9$ & $[-193.7,-134.3]$ & $-162.5$ & $[-198.4,-127.4]$\\
SFT-only base & $23.3$ & $[-12.8,71.4]$ & $24.3$ & $[-11.6,72.4]$\\
SFT-only / DPO single & $-9.3$ & $[-47.1,28.7]$ & $-2.2$ & $[-41.8,36.6]$\\
SFT-only / GRPO single & $-5.7$ & $[-41.5,28.9]$ & $-6.9$ & $[-43.1,29.1]$\\
SFT-only / DPO multi & $-0.8$ & $[-39.4,41.2]$ & $4.0$ & $[-34.9,46.2]$\\
SFT-only / GRPO multi & $17.5$ & $[-14.3,49.8]$ & $22.5$ & $[-10.5,54.6]$\\
\bottomrule
\end{tabular}
\end{table}

\section{Depletion assessment}
\label{si:depletion}

\subsection{Protocol}

We evaluate five fixed layouts: selected DPO, GRPO and variable-inventory-GA designs, and symmetric 16-Gd and 24-Gd references. The reflective two-dimensional geometry and beginning-of-life materials follow main Methods. Ordinary and Gd-bearing fuel pins receive separate depletable materials, with volumes defined for unit axial length.

OpenMC depletion couples transport to the 228-nuclide CASL PWR chain using the CE/CM integrator at 35~W\,gHM$^{-1}$ normalized to initial heavy-metal mass. The 27 burnup nodes are 0, 0.1, 0.25, 0.5, 1, 2, 3, 4, 5, 7.5 and 10 through 50 in increments of 2.5~MWd\,kgHM$^{-1}$, giving 26 intervals. Each transport calculation uses 20,000 particles per batch, 10 inactive batches and 20 active batches. The downward $k_\infty=1$ crossing is linearly interpolated between bracketing nodes.

At 50~MWd\,kgHM$^{-1}$, the residual-Gd effect is
\begin{equation}
\Delta k_{\mathrm{Gd}}=k_\infty^{\mathrm{endpoint,\ Gd\ removed}}-k_\infty^{\mathrm{endpoint}}.
\end{equation}
The perturbation removes every remaining Gd isotope while fixing all other endpoint nuclide densities and retaining the original depletion history. Independent original and perturbed transport variances are added to obtain its uncertainty.

\subsection{Results}

\begin{figure}[H]
\centering
\includegraphics[width=\textwidth,height=0.75\textheight,keepaspectratio]{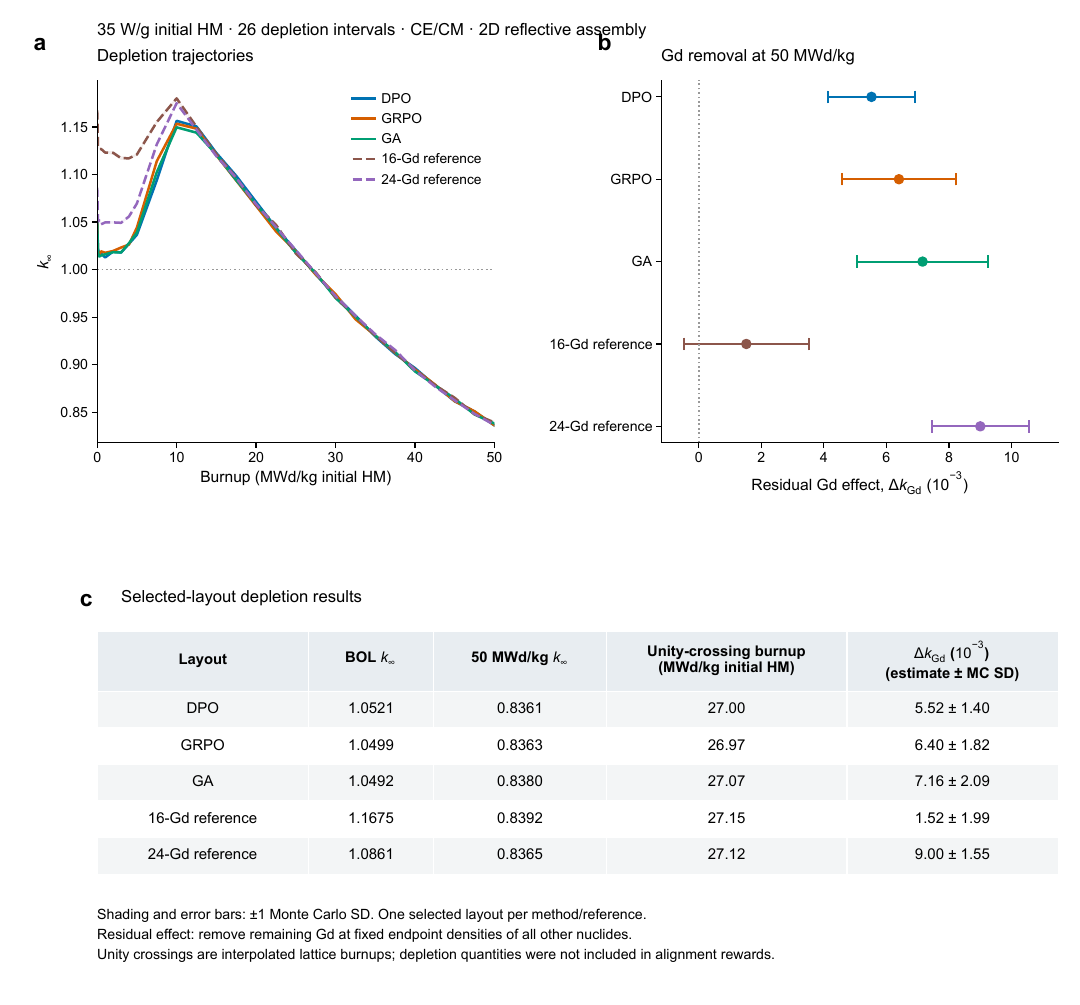}
\caption{\textbf{Depletion assessment beyond the alignment reward.} \textbf{a} Reflective-lattice multiplication factors for three selected optimized layouts and two constructed references over 26 CE/CM intervals to 50~MWd\,kgHM$^{-1}$ at 35~W\,gHM$^{-1}$. Shading denotes reported transport standard deviation. \textbf{b} Endpoint residual-Gd effect calculated by removing Gd while fixing other endpoint nuclide densities. Error bars denote propagated transport uncertainty. \textbf{c} Initial and endpoint multiplication factors, interpolated unity-crossing burnups and endpoint Gd effects. These diagnostics characterize the selected lattices under the stated depletion conditions.}
\label{si:depletion_figure}
\end{figure}

Supplementary Table~\ref{si:depletion_table} and Supplementary Fig.~\ref{si:depletion_figure} report all five layouts. Unity-crossing burnups lie between 26.97 and 27.15~MWd\,kgHM$^{-1}$. The aligned layouts' residual-Gd point estimates exceed that of the 16-Gd reference. These calculations evaluate one layout per method under fixed geometry and moderator conditions. Reported uncertainties cover transport sampling; depletion time-step convergence was not evaluated.

\begin{table}[htbp]
\centering
\small
\caption{\textbf{Depletion characteristics of five selected layouts.} Uncertainties in $\Delta k_{\mathrm{Gd}}$ are propagated transport standard deviations. Unity-crossing values are point estimates from the prescribed burnup grid, with depletion-discretization uncertainty unquantified.}
\label{si:depletion_table}
\begin{tabular}{lcccc}
\toprule
Layout & Initial $k_\infty$ & Endpoint $k_\infty$ & Unity crossing & $\Delta k_{\mathrm{Gd}}$\\
 & & & (MWd\,kgHM$^{-1}$) &\\
\midrule
DPO & 1.0521 & 0.8361 & 27.00 & $0.00552\pm0.00140$\\
GRPO & 1.0499 & 0.8363 & 26.97 & $0.00640\pm0.00182$\\
Variable GA & 1.0492 & 0.8380 & 27.07 & $0.00716\pm0.00209$\\
16-Gd reference & 1.1675 & 0.8392 & 27.15 & $0.00152\pm0.00199$\\
24-Gd reference & 1.0861 & 0.8365 & 27.12 & $0.00900\pm0.00155$\\
\bottomrule
\end{tabular}
\end{table}

\section{Complete chemistry results}
\label{si:chemistry}

Supplementary Table~\ref{si:chemistry_endpoints} reports all policy endpoints for both initialization arms and target schedules under the exact catalog evaluation in main Methods. Chemistry training uses batch size 128. Supplementary Table~\ref{si:chemistry_changes} reports changes paired to each policy's same-seed supervised base. All eight setting/initialization/update comparisons reduce error and increase off-support probability in every seed.

\begin{table}[htbp]
\centering
\small
\caption{\textbf{Complete chemistry policy endpoints.} Mean $\pm$ sample standard deviation across five seeds. Absolute yield error is in yield percentage points. All within-policy quantities are exact catalog expectations.}
\label{si:chemistry_endpoints}
\begin{tabular}{llcc}
\toprule
Setting / initialization & Policy & Absolute error & Off-support mass\\
\midrule
\multirow{3}{*}{Single / CPT$+$SFT}
& SFT base & $50.040\pm0.536$ & $0.287\pm0.042$\\
& DPO & $29.886\pm0.246$ & $0.701\pm0.023$\\
& GRPO & $29.771\pm0.166$ & $0.714\pm0.018$\\
\midrule
\multirow{3}{*}{Single / SFT only}
& SFT base & $50.929\pm0.728$ & $0.669\pm0.012$\\
& DPO & $30.530\pm0.725$ & $0.728\pm0.003$\\
& GRPO & $29.878\pm0.116$ & $0.731\pm0.002$\\
\midrule
\multirow{3}{*}{Multi / CPT$+$SFT}
& SFT base & $28.971\pm0.158$ & $0.036\pm0.005$\\
& DPO & $24.283\pm0.591$ & $0.080\pm0.023$\\
& GRPO & $24.110\pm0.321$ & $0.074\pm0.013$\\
\midrule
\multirow{3}{*}{Multi / SFT only}
& SFT base & $29.352\pm0.143$ & $0.073\pm0.011$\\
& DPO & $24.402\pm0.683$ & $0.152\pm0.021$\\
& GRPO & $24.252\pm0.485$ & $0.127\pm0.034$\\
\bottomrule
\end{tabular}
\end{table}

\begin{table}[htbp]
\centering
\small
\caption{\textbf{Paired chemistry changes from the supervised base.} Mean $\pm$ sample standard deviation of five within-seed changes. Every error change is negative and every off-support change positive at the seed level.}
\label{si:chemistry_changes}
\begin{tabular}{llcc}
\toprule
Setting / initialization & Update & Error change & Mass change\\
\midrule
Single / CPT$+$SFT & DPO & $-20.154\pm0.299$ & $+0.414\pm0.057$\\
Single / CPT$+$SFT & GRPO & $-20.269\pm0.444$ & $+0.427\pm0.025$\\
Single / SFT only & DPO & $-20.399\pm0.454$ & $+0.059\pm0.013$\\
Single / SFT only & GRPO & $-21.052\pm0.746$ & $+0.062\pm0.010$\\
\midrule
Multi / CPT$+$SFT & DPO & $-4.688\pm0.444$ & $+0.044\pm0.023$\\
Multi / CPT$+$SFT & GRPO & $-4.861\pm0.289$ & $+0.038\pm0.015$\\
Multi / SFT only & DPO & $-4.950\pm0.612$ & $+0.079\pm0.023$\\
Multi / SFT only & GRPO & $-5.100\pm0.476$ & $+0.055\pm0.033$\\
\bottomrule
\end{tabular}
\end{table}

Allowing all measured combinations lowers the attainable error from 24.628 to 20.577 for the single target and from 10.830 to 8.627 across the ten targets. Exact full-catalog uniform sampling gives errors 57.063 and 37.158 with off-support probability 0.749683. Anchor-only uniform sampling gives errors 55.427 and 36.809 with zero off-support probability.

\subsection{Data provenance}

The input is the published Buchwald--Hartwig dataset of Ahneman et al.\ \cite{ahneman2018predicting}, using the \texttt{Plates1-3} worksheet of \texttt{Dreher\_and\_Doyle\_input\_data.xlsx} distributed through \texttt{rxn4chemistry/rxn\_yields}. The CSV export preserves values, units and row multiplicity, with columns \texttt{Ligand, Additive, Base, Aryl halide, Output}. Source and export checksums are supplied in \texttt{data\_provenance.json}. Each target schedule has 20 completed jobs covering five seeds, two initialization arms and two online updates. The reaction table retains its upstream distribution terms.

\backmatter
\bibliography{sn-bibliography}